\documentclass[runningheads]{llncs}

\usepackage{eccv}

\usepackage{eccvabbrv}

\usepackage{graphicx}
\usepackage{booktabs}

\usepackage[accsupp]{axessibility}  

\usepackage{hyperref}

\hypersetup{
    colorlinks,
    citecolor=black,
    filecolor=black,
    linkcolor=black,
    urlcolor=black
}

\usepackage{orcidlink}
\usepackage{pifont}

\usepackage{marvosym}

\begin{document}

\title{Relightable 3D Gaussians: Realistic Point Cloud Relighting with BRDF Decomposition and Ray Tracing} 

\titlerunning{Relightable 3D Gaussians}

\author{Jian Gao\inst{1}$^*$ \and
Chun Gu\inst{2}$^*$ \and Youtian Lin\inst{1} \and Zhihao Li \inst{3} \and Hao Zhu\inst{1} \and Xun Cao\inst{1} \and \\ Li Zhang\inst{2}\textsuperscript{\Letter} \and Yao Yao\inst{1}\textsuperscript{\Letter}}

\authorrunning{J.~Gao et al.}

\institute{Nanjing University\and Fudan University\and Huawei Noah’s Ark Lab}

\maketitle

\renewcommand{\thefootnote}{}
\footnotetext{$^*$Equally contributed.}

\begin{abstract}
In this paper, we present a novel differentiable point-based rendering framework to achieve photo-realistic relighting. To make the reconstructed scene relightable, we enhance vanilla 3D Gaussians by associating extra properties, including normal vectors, BRDF parameters, and incident lighting from various directions. From a collection of multi-view images, the 3D scene is optimized through 3D Gaussian Splatting while BRDF and lighting are decomposed by physically based differentiable rendering. To produce plausible shadow effects in photo-realistic relighting, we introduce an innovative point-based ray tracing with the bounding volume hierarchies for efficient visibility pre-computation. Extensive experiments demonstrate our improved BRDF estimation, novel view synthesis and relighting results compared to state-of-the-art approaches. The proposed framework showcases the potential to revolutionize the mesh-based graphics pipeline with a point-based pipeline enabling editing, tracing, and relighting.
\keywords{3D Gaussian Splatting, Relighting, Point based rendering}
\end{abstract}
\section{Introduction}
\label{sec:intro}

Reconstructing 3D scenes from multi-view images for photo-realistic rendering is a fundamental problem at the intersection of computer vision and graphics. Recently, 3D Gaussian Splatting (3DGS)~\cite{kerbl20233d} has been proposed and has gained significant attention from the community. The method employs a set of 3D Gaussian points to represent a 3D scene and projects these points onto a designated view through a tile-based rasterization. Attributes of each 3D Gaussian point are then optimized through the point-based differentiable rendering. Notably, 3DGS achieves real-time rendering with quality comparable to or even surpassing previous state-of-the-art methods (e.g., Mip-NeRF~\cite{barron2021mip}), with a training speed on par with the most efficient Instant-NGP~\cite{muller2022instant}. However, the current 3DGS is unable to reconstruct a scene that can be relighted under different lighting conditions, making the method only applicable to the task of novel view synthesis. In addition, ray tracing, a crucial component in achieving realistic rendering, remains an unresolved challenge in point-based representation, limiting 3DGS to more dedicated rendering effects such as shadowing and light reflectance.

\begin{figure}[tb]
\centering

\begin{subfigure}[b]{0.49\linewidth}
    \includegraphics[width=\linewidth]{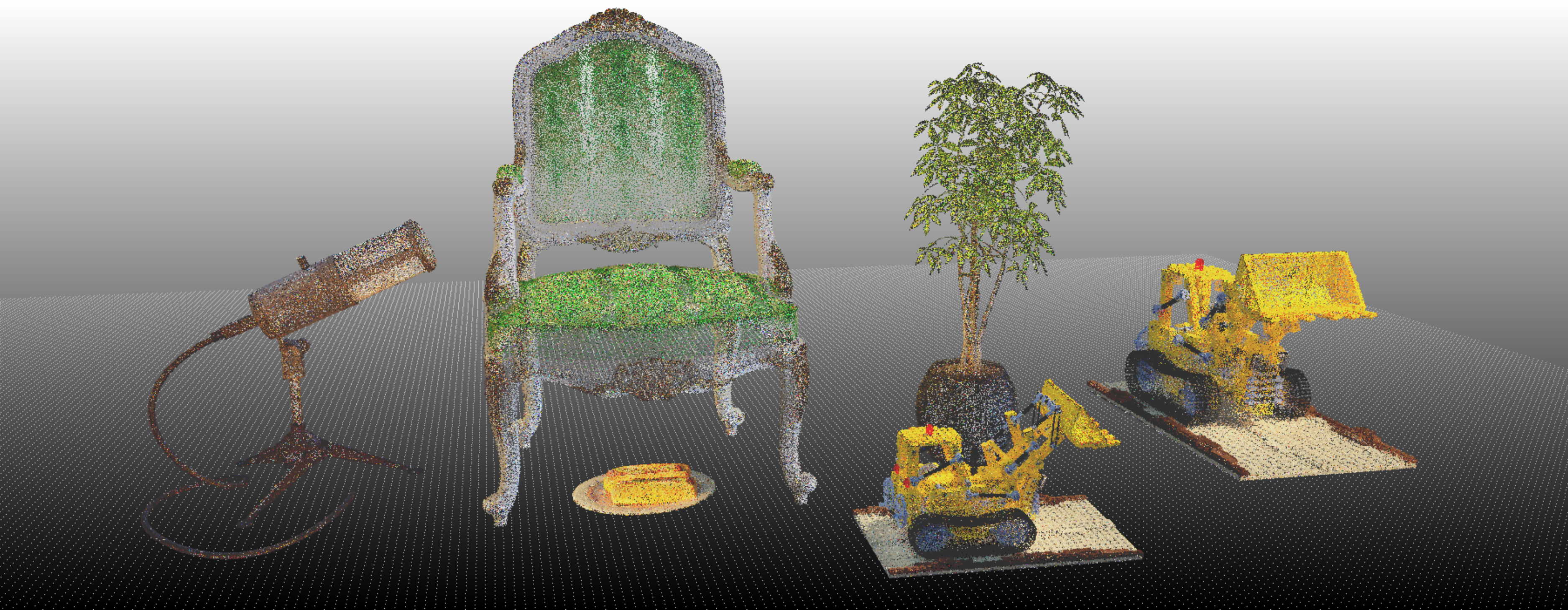}
    \caption{Point Cloud}
\end{subfigure} 
\begin{subfigure}[b]{0.49\linewidth}
    \includegraphics[width=\linewidth]{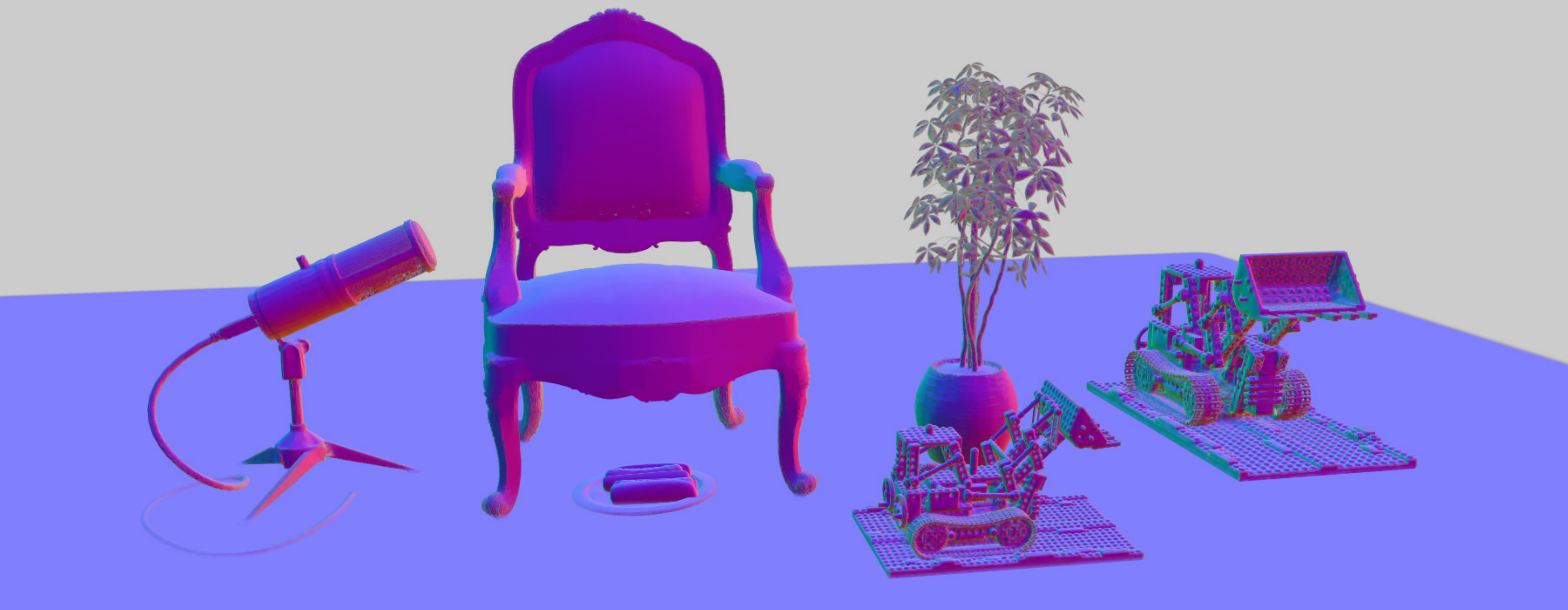}
    \caption{Normal}
\end{subfigure}
\begin{subfigure}[b]{0.49\linewidth}
    \includegraphics[width=\linewidth]{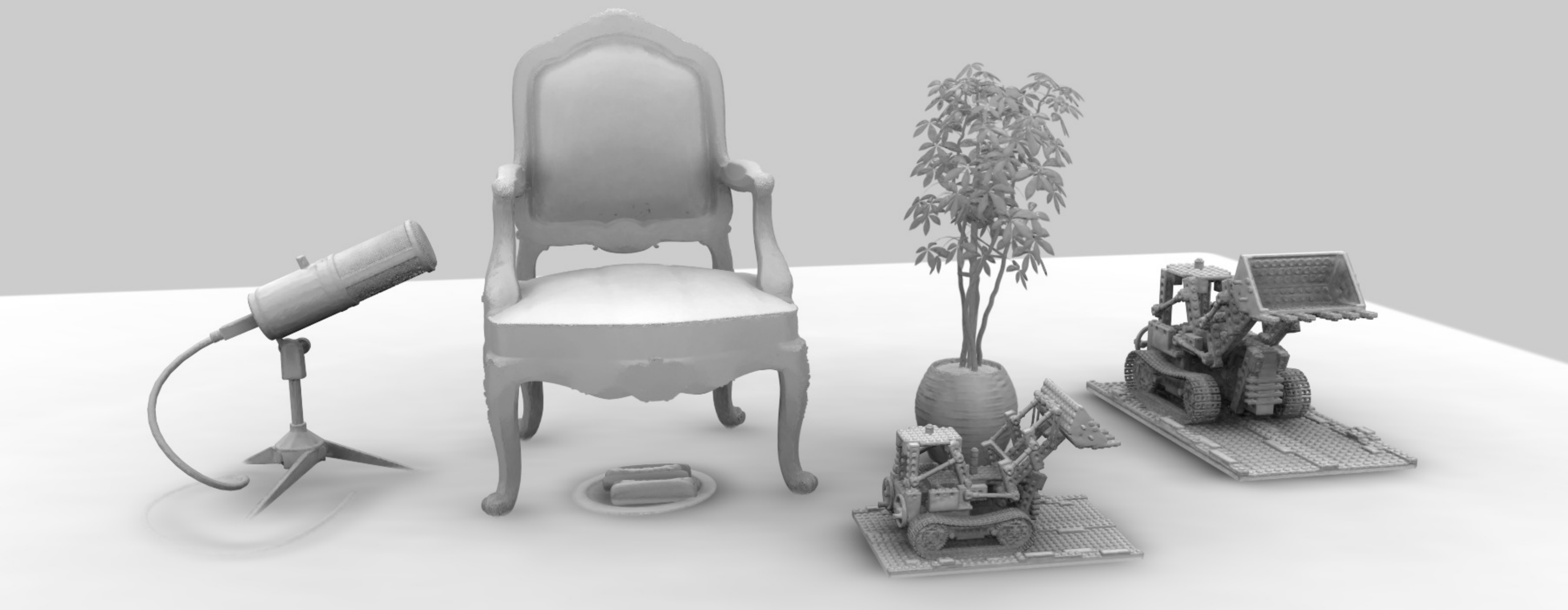}
    \caption{Ambient Occlusion}
\end{subfigure}
\begin{subfigure}[b]{0.49\linewidth}
    \includegraphics[width=\linewidth]{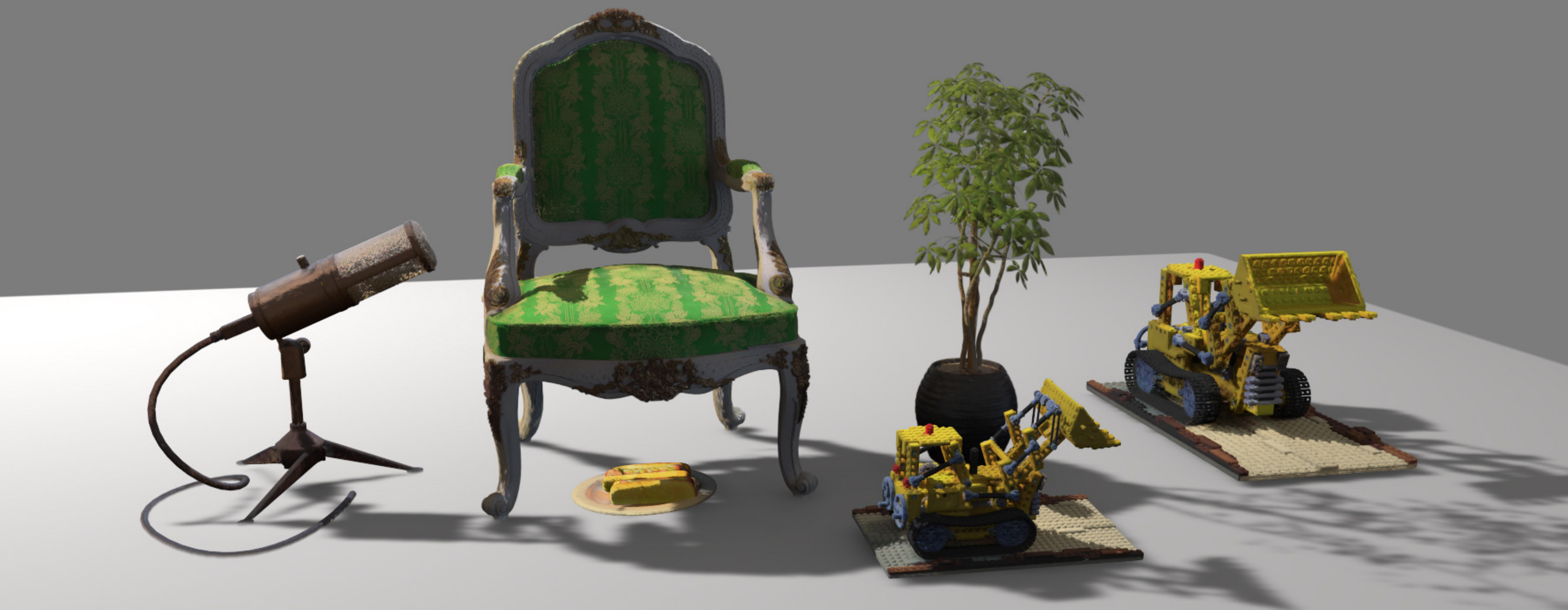}
    \caption{Physically Based Rendering}
\end{subfigure} 
\caption{\textbf{Visual results of our pipeline on a multi-object composition scene.} In our pipeline, we represent a scene as \textit{Relightable 3D Gaussians}. From multi-view images, we recover the geometry and materials of individual objects with \textit{inverse rendering} techniques (see Sec.~\ref{sec:optimization}). Then, objects are easily composited into a new scene, thanks to our explicit representation. After that, we solve the complex occlusions though \textit{point based ray tracing} (see Sec.~\ref{sec:raytracing}) and re-light the new scene. Ultimately, we achieve high-fidelity relighting with remarkably \textbf{realistic shadow}.
}
\label{fig:teaser}
\end{figure}

In this paper, we aim to reconstruct a relightable 3D Gaussian point cloud from multi-view images with a differentiable rendering framework and achieve realistic relighting. 
We make 3D Gaussian relightable by assigning it with additional normal, BRDF properties, and incident light information to model per-point light reflectance. 
In contrast to the plain alpha blending in the original 3DGS~\cite{kerbl20233d}, we apply physically based rendering (PBR) to get a PBR color for each 3D Gaussian point, which is then alpha-composited to obtain a rendered color for the corresponding image pixel. For robust material and lighting estimation, we split the incident light into a global environment map and an indirect incident light field. 
To capture accurate visibility for each 3D Gaussian, we propose a novel ray tracing method based on the bounding volume hierarchy (BVH), which enables efficient visibility pre-computing for real-time rendering. Additionally, the proposed ray tracing method can handle complex occlusion relationships in a novel multi-object composition scene, thus realizing realistic shadow effects.
Moreover, proper regularizations are introduced to enhance the geometry and mitigate the material-lighting ambiguity during the optimization, including constraints on depth distribution, smoothness priors, and a lighting regularization. 

Extensive experiments conducted across diverse datasets demonstrate the improved material and lighting estimation results and novel view rendering quality. Additionally, we illustrate the relightable and editable capabilities of our system through multi-object composition in a novel lighting environment. To summarize, major contributions of the paper include:
\begin{itemize}
\item We propose a material and lighting decomposition scheme for 3D Gaussian Splatting, where a normal vector, BRDF values, and incident lights are assigned and optimized for each 3D Gaussian point. 
\item We introduce a novel point-based ray tracing approach based on bounding volume hierarchy, enabling efficient visibility pre-computing for each 3D Gaussian point and rendering of a 3D scene with realistic shadow effect.
\item We demonstrate a comprehensive graphics pipeline solely based on a discretized point representation, supporting relighting, editing, and ray tracing of a reconstructed 3D point cloud.
\end{itemize}

\section{Related Works}
\label{sec:related}
\noindent{\bf Neural Radiance Field.}
Differentiable rendering techniques, exemplified by Neural Radiance Field (NeRF)~\cite{mildenhall2020nerf}, have garnered significant attention~\cite{barron2021mip,Chen2022ECCV, muller2022instant}. 
NeRF utilizes an implicit Multi-Layer Perceptron (MLP) that takes 3D positions and viewing directions as inputs to generate density and view-dependent colors for differentiable volume rendering. 
Despite its powerful neural implicit representation, both speed and quality can be improved. 
Efficiency improvements mainly focus on optimizing queries on neural fields and minimizing queries~\cite{reiser2021kilonerf, garbin2021fastnerf, hedman2021baking, hu2022efficientnerf, Chen2022ECCV, muller2022instant, fridovich2022plenoxels, chen2023mobilenerf}. 
Quality enhancements involve anti-aliasing~\cite{barron2021mip,barron2022mip,barron2023zip} or utilizing exterior supervision~\cite{deng2022depth,yu2022monosdf}, etc.

\noindent{\bf Differentiable Point Based Rendering.}
Using points as rendering primitives was first proposed in~\cite{levoy1985use}. To address the issue of resulting holey images when rasterizing discrete points directly, solutions fall into two categories. One approach directly employs points as rendering primitives and encodes the geometric and photometric features near the point using SIFT descriptors~\cite{pittaluga2019revealing} or neural descriptors~\cite{aliev2020neural, rakhimov2022npbg++}. This leads to a feature image with gaps, which is then decoded to recover a hole-free RGB image. The other category models a point as a primitive occupying a specific space, like a surfel~\cite{pfister2000surfels,yifan2019differentiable} or a 3D Gaussian~\cite{kerbl20233d}. The rendered image is then generated using a splatting technique. This technique saw significant development in the 2000s~\cite{zwicker2001surface, zwicker2002ewa, pfister2000surfels}, and in the era of differentiable rendering, PointRF~\cite{zhang2022differentiable}, DSS~\cite{yifan2019differentiable} and 3DGS~\cite{kerbl20233d} serve as notable examples. We assert that the point serves as a promising rendering primitive owing to its inherent ease of editing, as also corroborated in ~\cite{zhang2023frequency}.

\noindent{\bf Material and Lighting Estimation.}
Decomposing the materials and illumination of a scene from multi-view images presents a formidable challenge owing to its intrinsic high-dimensional complexity. Some methods simplified the problem under the assumption of a controllable light~\cite{azinovic2019inverse,guo2019relightables,park2020seeing,bi2020DRV,bi2020deep3d,nam2018practical,schmitt2020joint,bi2020NRF}. Subsequent research explores more complex lighting models to cope with realistic scenarios. 
NeRV~\cite{srinivasan2021nerv} and PhySG~\cite{zhang2021physg} leverage an environmental map to manage arbitrary lighting conditions. NeRD~\cite{boss2021nerd} addresses the challenge posed by images captured under varying illumination by attributing Spherical Gaussians to each image. IRON~\cite{zhang2022iron} introduces an innovative edge sampling algorithm tailored for neural SDFs. NeRFactor~\cite{zhang2021nerfactor} exploits light visibility to achieve superior material and lighting decomposition. Further studies~\cite{zhang2022differentiable,hasselgren2022shape} address the consideration of indirect lighting, leading to enhanced BRDF estimation quality. 
\cite{munkberg2022extracting,hasselgren2022nvdiffrecmc} utilize differentiable marching tetrahedrons for direct optimization on mesh surfaces. ~\cite{boss2022samurai,kuang2022neroic} deal with varying cameras, illumination, and backgrounds. Ref-NeRF~\cite{verbin2022ref} suggests the employment of integrated direction encoding to ameliorate the reconstruction fidelity of reflective objects. \cite{munkberg2022extracting, liu2023nero} use split-sum approximation to approximate the shading effects. 
NeMF~\cite{zhang2023nemf} represents the scene as a microflake volume. NeFII~\cite{wu2023nefii} integrates lights through path tracing with Monte Carlo sampling. 
InvRender~\cite{zhang2022modeling} computes indirect illumination by directly leveraging the neural radiance field, rather than concurrently estimating it alongside the decomposition of direct lighting and materials. TensoIR~\cite{jin2023tensoir} performs inverse rendering based on tensor factorization and neural fields. NeILF~\cite{yao2022neilf} and NeILF++~\cite{zhang2023neilfpp} expresses the incident lights as a neural incident light field, while NeILF++ integrates VolSDF~\cite{yariv2021volume} with NeILF through inter-reflection. However, existing schemes rarely delve into BRDF estimation for point clouds, and the estimation and rendering quality remain to be improved.
\section{Relightable 3D Gaussians}
\label{sec:optimization}
\begin{figure}[tb] 
  \includegraphics[width=\linewidth]{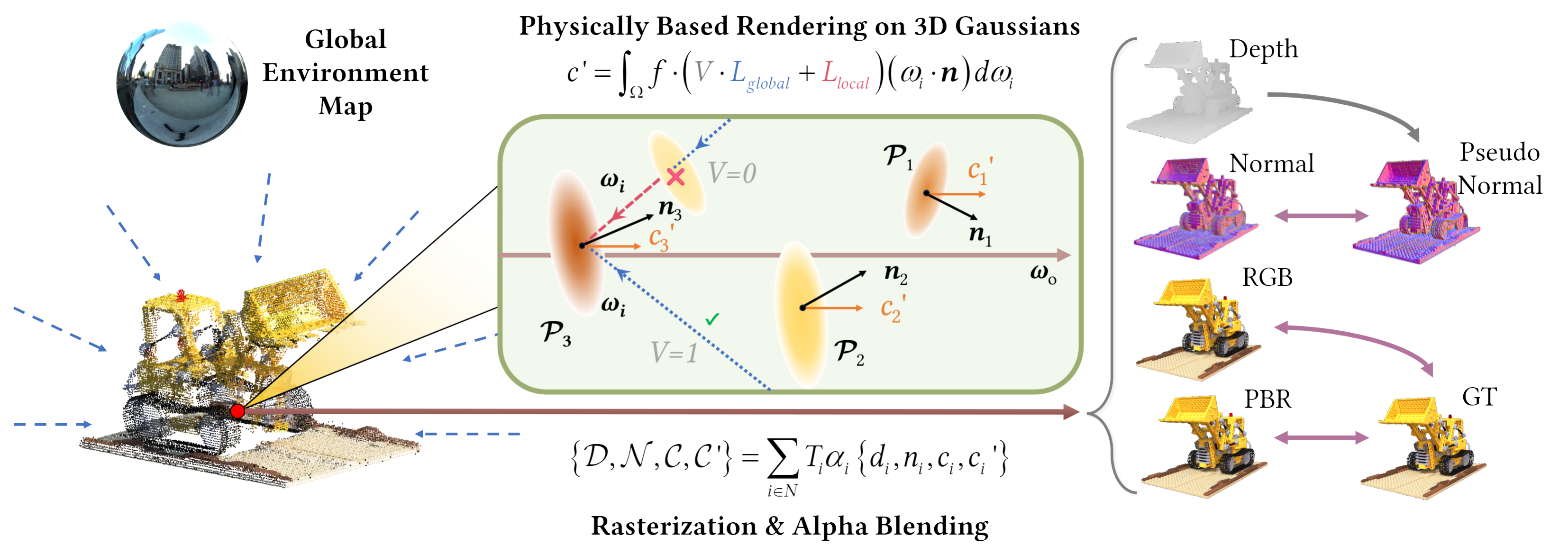}
  \caption{\textbf{The proposed differentiable rendering pipeline.} Starting with a collection of 3D Gaussians that embody geometry, materials, and lighting attributes, we first execute the physically based rendering equation for each 3D Gaussian to determine its outgoing radiance, denoted as $\boldsymbol{c'}$. Following this, we perform rasterization and alpha blending to obtain vanilla color map $\mathcal{C}$, PBR color map $\mathcal{C'}$, depth map $\mathcal{D}$, normal map $\mathcal{N}$, etc. To optimize relightable 3D Gaussians, we utilize the ground truth image $\mathcal{C}_{gt}$ and the pseudo normal map derived from $\mathcal{D}$ for supervision.} 
  \label{fig:pipeline}
\end{figure}

In this section, we introduce a novel pipeline to decompose materials and lighting from multi-view images based on 3D Gaussian Splatting (3DGS)~\cite{kerbl20233d}. An overview of our pipeline is shown in Fig.~\ref{fig:pipeline}. 
\subsection{Preliminary}
\label{sec:preliminaries}
Distinct from the widely adopted Neural Radiance Field, 3DGS~\cite{kerbl20233d} employs explicit 3D Gaussian points as its rendering primitives. 
A 3D Gaussian point is mathematically defined as:
\begin{equation}
G(\boldsymbol{x}) = exp(-\frac{1}{2}(\boldsymbol{x}-\boldsymbol{\mu})^\top\Sigma^{-1}(\boldsymbol{x}-\boldsymbol{\mu})) ,
\label{eq:3d_gaussian}
\end{equation}
where $\boldsymbol{\mu}$ and $\Sigma$ denote the 3D spatial mean and covariance matrix, respectively. Each Gaussian is also equipped with an opacity $o$ and a view-dependent color $\boldsymbol{c}$.

The rendering process in 3DGS is divided into two main steps. Firstly, 3D Gaussians are projected to 2D Gaussians on the image plane. The 2D means are determined by accurate projection of 3D means, while the 2D covariance matrices are approximated by:
$\Sigma'=\boldsymbol{J}\boldsymbol{W}\Sigma\boldsymbol{W}^\top\boldsymbol{J}^\top$,
where $\boldsymbol{W}$ and $\boldsymbol{J}$ denote the viewing transformation and the Jacobian of the affine approximation of perspective projection transformation~\cite{zwicker2001surface}. 
Subsequently, the pixel color is derived by alpha blending $N$ ordered 2D Gaussians from front to back:
\begin{equation}
\mathcal{C}=\sum_{i\in{N}}T_{i}\alpha_{i}\boldsymbol{c}_{i}, \hspace{0.5em} T_i = \prod_{j=1}^{i-1}(1-\alpha_{j}) .
    \label{eq:alpha_blending}
\end{equation}
Here, $\alpha$ is obtained by multiplying the opacity $o$ with the 2D covariance's contribution computed from $\Sigma'$ and pixel coordinate in image space~\cite{kerbl20233d}.
In implementation details, the covariance matrix $\Sigma$ is parameterized as a unit quaternion $\boldsymbol{q}$ and a 3D scaling vector $\boldsymbol{s}$ to maintain its meaningful interpretation throughout optimization. Additionally, view-dependent color $\boldsymbol{c}_{i}$ is represented through a set of Spherical Harmonics (SH).

In summation, a 3D scene is represented by a collection of 3D Gaussians, with the $i^{th}$ Gaussian $\mathcal{P}_i$ parameterized as $\{\boldsymbol{\mu}_i, \boldsymbol{q}_i, \boldsymbol{s}_i, o_i, \boldsymbol{c}_i\}$. 

\subsection{Geometry Enhancement}
\label{sec:geometry enhancement}
Scene geometry, specifically surface normal, is essential for realistic physically based rendering, which will be discussed in Sec.~\ref{sec:brdf light}.

\noindent{\bf Normal Estimation.}
We incorporate a normal attribute $\boldsymbol{n}$ for each 3D Gaussian and try to solve it. A potential methodology treats the spatial mean of a 3D Gaussian as a conventional point, and executes normal estimation based on the local planar assumption. However, this approach cannot provide accurate normal estimation. The reasons are two fold: First, the reconstructed Gaussian point cloud is often sparse. More critically, the Gaussian points are naturally \textit{soft}, which means these points are not precisely aligned with the object surface.

To address these limitations, we propose to optimize $\boldsymbol{n}$ from an initial random vector via back-propagation for each 3D Gaussian. We perceive the depth of all Gaussians along a ray as a distribution, and estimate the pixel depth as the expectation of this distribution. Similarly, we determine the normal for each pixel. This process is described by:
\begin{equation}
    \{\mathcal{D}, \mathcal{N}\} = \sum_{i\in N} w_i\{d_{i},\boldsymbol{n}_{i}\},
    \label{eq:render_depth}
\end{equation}
where $d_{i}$, $\boldsymbol{n}_{i}$ and $w_i={T_{i}\alpha_{i}}/ \sum_{i\in{N}}T_{i}\alpha_{i}$  denote the depth, normal and weight of the point. We then encourage the consistency between the rendered normal $\mathcal{N}$ and the pseudo normal $\tilde{\mathcal{N}}$, which is computed from the rendered depth $\mathcal{D}$ under the local planarity assumption. The normal consistency is quantified as follows:
\begin{equation}
    \mathcal{L}_{n}=\|\mathcal{N}-\tilde{\mathcal{N}}\|_2 .
    \label{eq:cost_normal}
\end{equation}

\noindent{\bf Normal Gradient Based Densification.}
To achieve superior rendering fidelity in detail regions, vanilla 3DGS~\cite{kerbl20233d} densifies 3D Gaussians through the gradient of view space points. Drawing upon this, to improve normal recovery in thin regions, we introduce an additional densification criterion on the gradient of normals. Specifically, we densify Gaussians whose normal gradient exceed a threshold $T_{\boldsymbol{n}}$.

\noindent{\bf Constraint on Depth Distribution.}
Given our assumption of the object possessing an accurate surface, we enforce a constraint on the depth distribution by minimizing the uncertainty. The uncertainty is defined as:

\begin{equation}
    \mathcal{L}_{u} = \mathcal{D}_{sq} - \mathcal{D}^2,
    \label{eq:depth uncertainty}
\end{equation}
where $\mathcal{D}_{sq}=\sum_{i\in N}w_id_i^2$, and $\mathcal{D}$ is the depth estimation defined in Eq.~\ref{eq:render_depth}.
This constraint on the depth distribution drives Gaussian points to the object surface, thereby improving geometric reconstruction.

\noindent{\bf Object Mask Constraint.}
If there is a mask indicating the object, we can constrain the optimization by a binary cross entropy loss~\cite{wang2021neus}:
\begin{equation}
\begin{split}
    \mathcal{L}_{O} = -M\log{O} - (1-M)\log{(1-O)},
\end{split} 
\end{equation}
where $M$ is the object mask and $O=\sum_{i\in{N}}T_{i}\alpha_{i}$ . With this constraint, $O$ is forced to be aligned with the distribution of $M$, and so we get opaque surface. 

\subsection{BRDF and Light Modeling}
\label{sec:brdf light}
\noindent{\bf Rendering Equation.}
We use the the rendering equation~\cite{kajiya1986rendering} to model light interaction with surfaces, accounting for their material properties and geometry. It is given by:

\begin{equation}
L_{o}(\boldsymbol{\omega_{o}}, \boldsymbol{x}) = \int_{\Omega}f(\boldsymbol{\omega_{o}}, \boldsymbol{\omega_{i}}, \boldsymbol{x})L_{i}(\boldsymbol{\omega_{i}}, \boldsymbol{x})(\boldsymbol{\omega_{i}}\cdot\boldsymbol{n})d\boldsymbol{\omega_{i}}
\label{eq:rendering_equation},
\end{equation}
where $\boldsymbol{x}$ and $\boldsymbol{n}$ are the surface point and its normal vector, $f$ is the Bidirectional Reflectance Distribution Function (BRDF), and $L_{i}$ and $L_{o}$ denote the incoming and outgoing radiance in directions $\boldsymbol{\omega_{i}}$ and $\boldsymbol{\omega_{o}}$. $\Omega$ signifies the hemispherical domain above the surface.

Prior methods~\cite{zhang2021physg,zhang2023neilfpp} typically begin with the acquisition of intersection points between rays and surface through differentiable rendering. Then they apply the rendering equation at these points to facilitate Physical Based Rendering (PBR). However, these approach present significant challenges in the point-based framework. Certainly, it is feasible to extract surfaces in 3DGS~\cite{guedon2023sugar}; but, it necessitates a coordinate-based incident light field for inverse PBR. The querying of this field at millions of points during every iteration imposes a substantial computational burden. Moreover, it is pertinent to acknowledge that accurately extracting geometric surfaces presents considerable challenges in 3DGS system.

To tackle this issue, we propose to compute PBR color $\{c_i'\}_{i=0}^{N}$ for each 3D Gaussian, and then obtain the PBR image $\mathcal{C'}$ through alpha-blending, as shown in Fig.~\ref{fig:pipeline}.
This method is more efficient for two main reasons. First, PBR is performed on fewer 3D Gaussians rather than image pixels in all input views, as each Gaussian affects multiple pixels based on its scale. Second, by allocating discrete attributes for each Gaussian, we avoid the need for global neural fields.

\noindent{\bf BRDF Parameterization.}
To make 3D Gaussians relightable, we assign BRDF properties to each Gaussian and adopt a simplified Disney BRDF model~\cite{burley2012physically}. The BRDF properties include an albedo $\boldsymbol{b}\in[0, 1]^3$ and a roughness $r\in[0, 1]$, and the BRDF $f$ in Eq. \ref{eq:rendering_equation} is divided into a diffuse term $f_{d} = \frac{\boldsymbol{b}}{\pi}$ and a specular term:
\begin{equation}
    f_{s}(\boldsymbol{\omega}_{o}, \boldsymbol{\omega}_{i})=\frac{D(\boldsymbol{h};r) \cdot F(\boldsymbol{\omega}_{o},\boldsymbol{h}) \cdot G(\boldsymbol{\omega}_{i},\boldsymbol{\omega}_{o}, h; r)}{(\boldsymbol{n} \cdot \boldsymbol{\omega}_{i}) \cdot (\boldsymbol{n} \cdot \boldsymbol{\omega}_{o})},
    \label{eq:specular_term}
\end{equation} where $\boldsymbol{h}=({\boldsymbol{\omega}_{i}+\boldsymbol{\omega}_{o}}) / 2$ is the half vector, \textit{D}, \textit{F} and \textit{G} denote the normal distribution function, Fresnel term and geometry term. 
 
\noindent{\bf Incident Light Modeling.}
Optimizing a NeILF for each Gaussian can be excessively unconstrained, leading to challenges in accurately decomposing incident lights from its appearance. To address this issue, we apply a prior by partitioning the incident light into globally shared direct component and individual per-Gaussian indirect components. The sampled incident light at a Gaussian from direction $\boldsymbol{\omega}_{i}$ is represented as:
\begin{equation}
    L_i(\boldsymbol{\omega}_{i}) = V(\boldsymbol{\omega}_{i})\cdot L_{direct}(\boldsymbol{\omega}_{i}) + L_{indirect}(\boldsymbol{\omega}_{i}) ,
    \label{eq:light_representation}
\end{equation}
where $V(\boldsymbol{\omega}_{i})$ represents the visibility term which will be further discussed in Sec.~\ref{sec:raytracing}, the indirect light term  $L_{indirect}$ is parameterized by 3-level SH, denoted as $\boldsymbol{l}$, while the direct light term $L_{direct}$ is parameterized as a globally shared 16x32 environment map, denoted as $\boldsymbol{l}^{env}$. Despite the utilization of relatively low-level Spherical Harmonics (SH), we can still capture some inter-reflection effects, as the rendering color of a given pixel arises from the collaboration of multiple relightable 3D Gaussians.

For each 3D Gaussian, we sample $N_s$ incident directions over the hemisphere space through Fibonacci sampling~\cite{yao2022neilf} to provide numerical integration for Eq.~\ref{eq:rendering_equation}. The PBR color of each Gaussian is then given by:
\begin{equation}
    \boldsymbol{c'(\boldsymbol{\omega}_{o})} = \sum_{i=0}^{N_s}(f_d+f_s(\boldsymbol{\omega}_{o}, \boldsymbol{\omega}_{i}))L_{i}(\boldsymbol{\omega}_{i})(\boldsymbol{\omega}_{i}\cdot\boldsymbol{n})\Delta\boldsymbol{\omega}_{i},
    \label{eq:pbr rendering}
\end{equation}
where $\boldsymbol{\omega}_{i}$ is the solid angle.

To summarize, our method represents a 3D scene as a set of relightable 3D Gaussians and a global environment light $\boldsymbol{l}^{env}$, where the $i^{th}$ Gaussian $\mathcal{P}_i$ is parameterized as $\{\boldsymbol{\mu}_i, \boldsymbol{q}_i, \boldsymbol{s}_i, o_i, \boldsymbol{c}_i, \boldsymbol{n}_i, \boldsymbol{b}_i, r_i, \boldsymbol{l}_i\}$. 

\subsection{Regularizations}
\label{sec:regularization}

To mitigate the materials-lighting ambiguity~\cite{yao2022neilf}, proper regularizations are utilized to facilitate their plausible decomposition.

\noindent{\bf Light Regularization.}
We apply a light regularization assuming a near-natural white incident light~\cite{liu2023nero}:
\begin{equation}
\label{eq:reg_light}
    \mathcal{L}_{light} = \sum\nolimits_{c}(L_{c} - \frac{1}{3}\sum\nolimits_{c}L_{c}), c\in\{R,G,B\}.
\end{equation}

\noindent{\bf Smoothness Priors.}
We expect that the BRDF properties do not change drastically in homogeneous areas~\cite{yao2022neilf}. We define a smooth constraint on roughness as:
\begin{equation}
\label{eq:reg_smooth}
    \mathcal{L}_{s, r} = \Vert \nabla R \Vert \exp(-\Vert \nabla C_{gt} \Vert) ,
\end{equation}
where $R=\sum_{i\in{N}}w_ir_{i}$ is the rendered roughness map. Similarly, we also define smoothness constraints $\mathcal{L}_{s, n}$ and $\mathcal{L}_{s, b}$ on normal and albedo.

\section{Point-based Ray Tracing}
\label{sec:raytracing}

For real-time realistic rendering with plausible shadow effects on relightable 3D Gaussians, we introduce a novel point-based ray tracing approach in this section. 

\subsection{Ray Tracing on 3D Gaussians}
\label{sec:3DG_raytracing}
Our proposed ray tracing technique on 3D Gaussians is built upon the Bounding Volume Hierarchy (BVH), enabling efficient visibility querying along a ray. Our method adopts the idea from ~\cite{karras2012maximizing}, an in-place algorithm for constructing a binary radix tree that maximizes parallelism and facilitates real-time BVH construction. 
Specifically, we construct a binary radix tree from a given set of 3D Gaussians, where each leaf node represents the tight bounding box of a Gaussian, and each internal node denotes the bounding box of its two children.

\begin{figure}[t]
    \centering
    \begin{subfigure}{.23\textwidth}
        \centering
        \includegraphics[width=\linewidth]{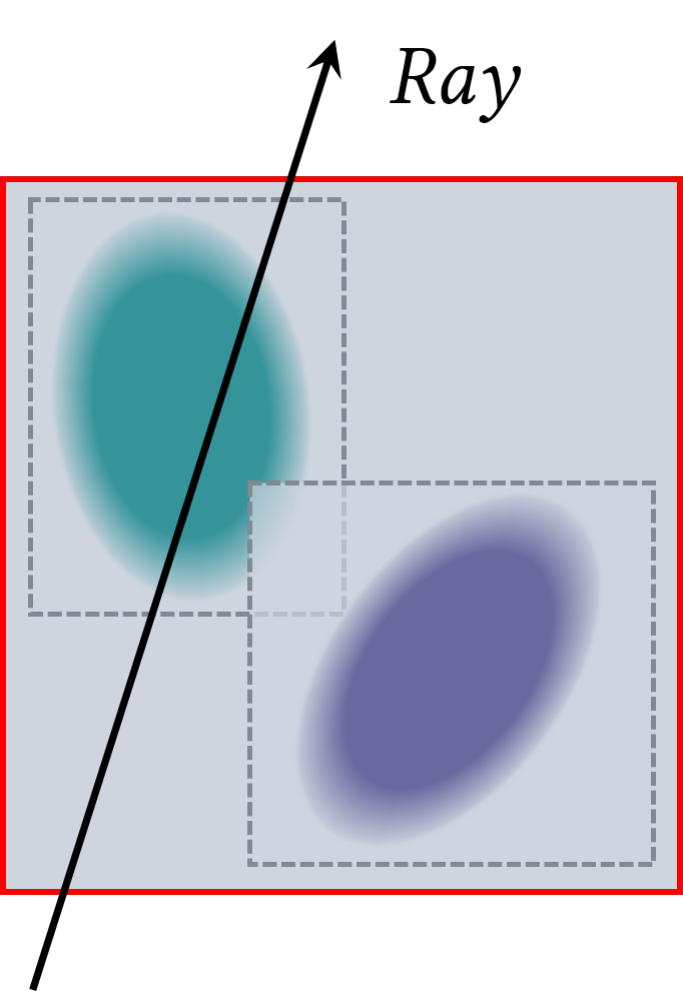}
    \caption{Root Node}
    \label{fig:trace_1}
    \end{subfigure}
    \hfill
    \begin{subfigure}{.23\textwidth}
        \centering
        \includegraphics[width=\linewidth]{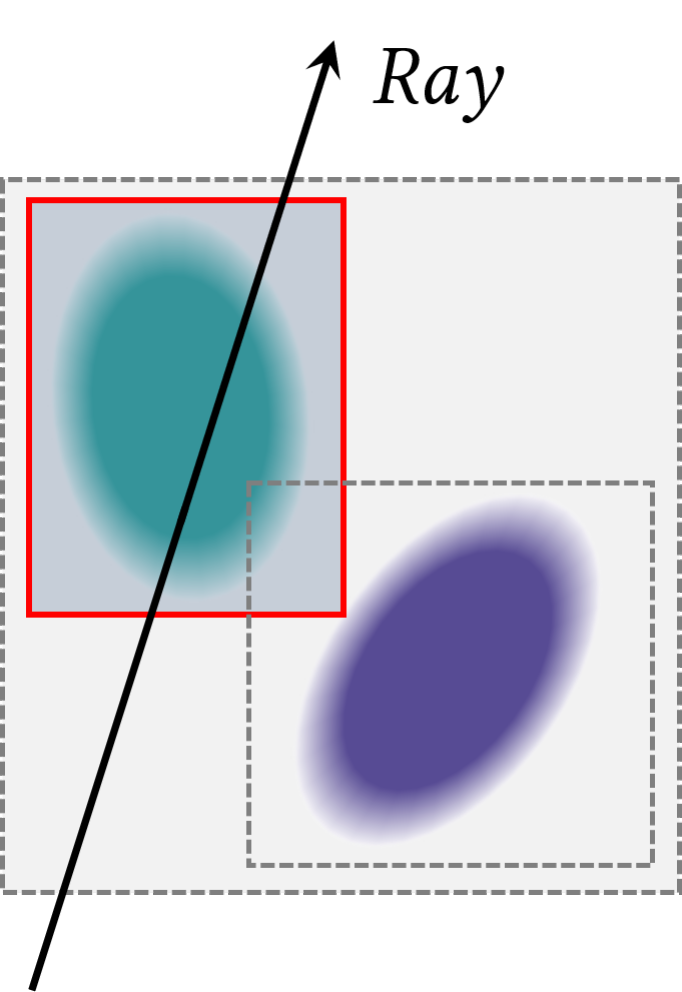}
    \caption{Child Node}
    \label{fig:trace_2}
    \end{subfigure}
    \hfill
    \begin{subfigure}{.23\textwidth}
        \centering
        \includegraphics[width=\linewidth]{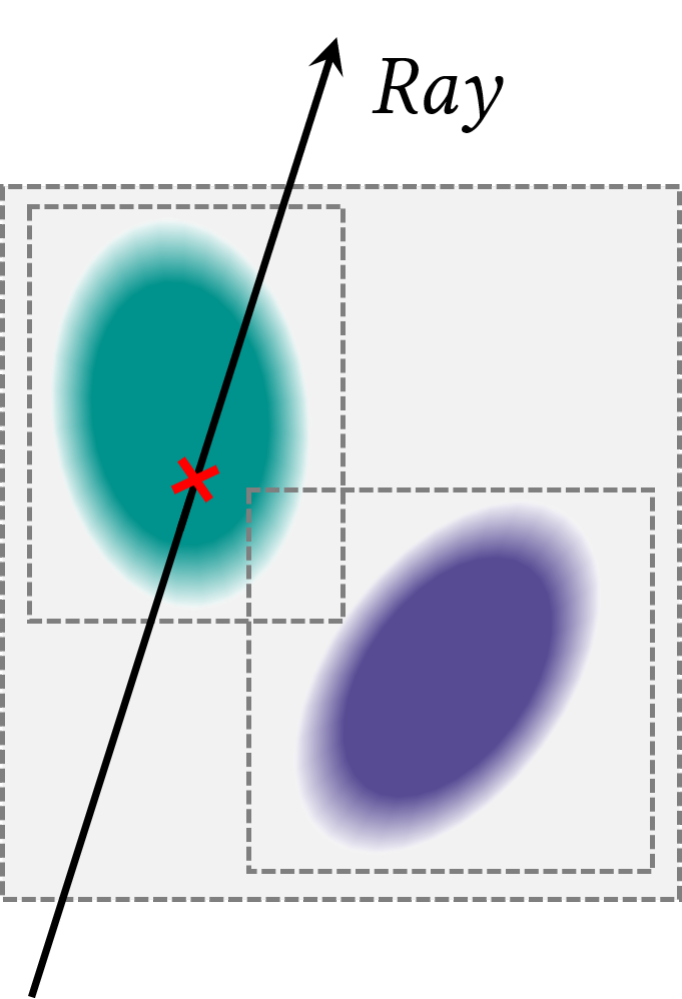}
        \caption{Gaussian}
    \label{fig:trace_3}
    \end{subfigure}
    \hfill
    \begin{subfigure}{.23\textwidth}
        \centering
        \includegraphics[width=\linewidth]{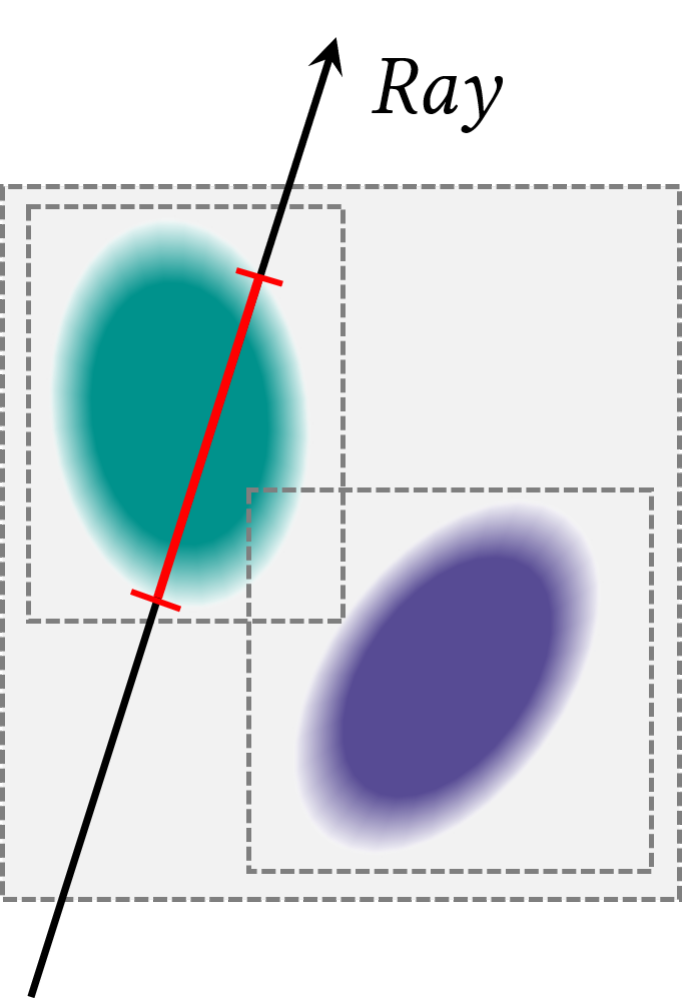}
        \caption{Gaussian}
    \label{fig:trace_4}
    \end{subfigure}
\caption{\textbf{Intersection tests in point-based ray tracing.} Intersection point between ray and Gaussian is obtained by three steps: (a) intersect the BVH root node; (b) dive into the intersected child nodes recursively until the leaf node; (c) perform Eq.~\ref{eq:intersect} to get the equivalent intersection point. (d) shows that a 3D Gaussian actually has non-negligible effect on a segment of a ray.} 
\label{fig:trace}
\end{figure} 

Unlike ray tracing with opaque polygonal meshes, where only the closest intersection point is required, ray tracing with semi-transparent Gaussians necessitates accounting for all Gaussians potentially influencing the ray's transmittance. The process of ray tracing on 3D Gaussian points can be described as follows: the ray travels from the camera center and accumulates transmittance as it passes through 3D Gaussian points until the transmittance is zero. The first key issue to be addressed is how to define the contribution of a single 3D Gaussian to the transmittance. 
As discussed in Sec.\ref{sec:preliminaries}, transforming a 3D Gaussian to a 2D Gaussian involves an approximation, which complicating the identification of accurate intersection between the Gaussian and a ray. As shown in Fig.~\ref{fig:trace_4}, a 3D Gaussian actually has non-negligible effect on a segment of a ray. 
Thus, drawing inspiration from~\cite{keselman2023flexible}, we approximate the intersection of ray with 3D Gaussian as a point where the 3D Gaussian's contribution peaks, as showed in Fig. 3(3). The equivalent intersection point is defined as:

\begin{equation}
\boldsymbol{r_x} = \boldsymbol{r_o}+t_j\boldsymbol{r_d},
\label{eq:intersect}
\end{equation}
where \(\boldsymbol{r_o}\) denotes the origin, \(\boldsymbol{r_d}\) is direction of the ray, corresponding to the previously mentioned $\boldsymbol{\omega}_{i}$, $t_j$ is defined as: 
\begin{equation}
    t_j = \frac{(\boldsymbol{\mu} - \boldsymbol{r_o})^T \boldsymbol{\Sigma} \boldsymbol{r_d}}{\boldsymbol{r_d}^T \boldsymbol{\Sigma} \boldsymbol{r_d}}.
    \label{eq:intersect2}
\end{equation}
Subsequently, we approximate the contribution of this 3D Gaussian to the ray as $\alpha_j$ at the equivalent
intersection point $\boldsymbol{r_x}$.

Considering the transmittance along a ray: $T_i = \prod_{j=1}^{i-1}(1-\alpha_{j})$, it is evident that the order of \(\alpha_{j}\) does not affect \(T_i\). In other word, the order in which Gaussians are encountered along a ray does not impact the overall transmittance.
As illustrated in Fig.~\ref{fig:trace}, starting from the root node of the binary radix tree, intersection tests are recursively performed between the ray and the bounding volumes of each node's children. 
Upon reaching a leaf node, the associated Gaussian is identified. Through this traversal, the transmittance \( T \) is progressively attenuated:
\begin{equation}
T_i = (1 - \alpha_{i-1})T_{i-1}, \quad \text{for } i = 1, \ldots, j-1 \text{ with } T_1 = 1.
\end{equation}
To speed up ray tracing, the process is early terminated if a ray's transmittance drops below a certain threshold $T_{min}$.

\subsection{Visibility Pre-computation}
\label{sec:est_visibility}
In Sec.~\ref{sec:brdf light}, a potential ambiguity in the decomposition of indirect and direct lights has been discerned. While the essential visibility term can be derived via the proposed ray tracing on 3D Gaussians (Sec.~\ref{sec:3DG_raytracing}), querying the visibility term through ray tracing online proves challenging due to the computational complexity. Nevertheless, given our exclusive focus on static scenes, we can pre-compute the visibility term $V(\boldsymbol{\omega}_{i})$ across the hemispherical domain determined by $\boldsymbol{n}$ for each Gaussian, and subsequently integrate it into the rendering equation.

Hence, we divide the optimization process into two stages. In the first stage, we optimize a vanilla 3D Gaussian point cloud, augmented with an additional normal vector $\boldsymbol{n}$ for each Gaussian. Immediately afterwards, we pre-compute the per-Gaussian visibility through the proposed ray tracing. In the second stage, we lock the geometry of 3D Gaussians and focus solely on optimizing the material and lighting parameters using the pipeline described in Fig.~\ref{fig:pipeline}.

\subsection{Point Based Relighting Pipeline}
\label{sec:3DG_relighting}
Based on our relightable 3D Gaussians, we devise a point based graphics pipeline that seamlessly integrates \textbf{effortless scene editing} and \textbf{realistic relighting}. To our knowledge, there is currently no point-based rendering pipeline that effectively accomplishes both tasks. Although explicit point representations make scene editing easy, achieving photo-realistic relighting proved nearly insurmountable in existing point-based rendering pipeline. Conversely, in inverse rendering approaches based on implicit representations, while highly realistic relighting is feasible, scene editing presents a challenging endeavor.
 
As an illustration, we concentrate on relighting in a multi-object composition scene. Initially, our pipeline computes the visibility $V(\boldsymbol{\omega}_{i})$ for each Gaussian point via ray tracing (Sec.~\ref{sec:3DG_raytracing}). Despite the challenging inter-object occlusions introduced by composition, our proposed point-based ray tracing method adeptly manages these complexities and ensures accurate occlusion updates within the novel scene. Subsequently, the rendering process unfolds (Sec.~\ref{sec:brdf light}), starting with Gaussian-level PBR and ending with alpha blending. Throughout this process, the original indirect lighting of each object is discarded. Consequently, we achieve relighting with remarkably vivid shadow effects solely based on a discrete set of points, as illustrated in Fig.~\ref{fig:teaser}.
\section{Experiments}
\label{sec:experiments}
\subsection{Training Details}
\label{sec:details}

As pointed out in Sec.~\ref{sec:est_visibility}, the training procedure is divided into two stages. The first stage follows the setting of 3DGS~\cite{kerbl20233d}, along with the proposed normal gradient-based densification (Sec. ~\ref{sec:geometry enhancement}) where $T_{\textbf{n}}$ is set as $2\times 10^{-9}$. In the second stage, we sample $N_s=64$ incident rays per Gaussian point for PBR. We train our model for 30,000 iterations in the initial stage and 10,000 iterations in the second stage. 
The loss in the first stage is calculated by:
\begin{equation}
\label{eq:loss_stage1}
    \mathcal{L} = \lambda_{1} \mathcal{L}_{1} + \lambda_{ssim} \mathcal{L}_{ssim} + \lambda_{n} \mathcal{L}_{n} + \lambda_{s,n} \mathcal{L}_{s,n} + \lambda_{O} \mathcal{L}_{O} + \lambda_{u} \mathcal{L}_{u},
\end{equation}
where $\{\lambda_{1}, \lambda_{ssim}, \lambda_{n}, \lambda_{s,n}, \lambda_{O}, \lambda_{u}\}$ are set to $\{0.8, 0.2, 0.01, 0.01, 0.01, 0.01\}$, respectively. The loss in the second stage is given by:
\begin{equation}
\label{eq:loss_stage2}
    \mathcal{L} = \lambda_{1} \mathcal{L}_{1} + \lambda_{ssim} \mathcal{L}_{ssim} + \lambda_{l} \mathcal{L}_{l} + \lambda_{s,b} \mathcal{L}_{s,b} + \lambda_{s, r} \mathcal{L}_{s, r},
\end{equation}
where $\{\lambda_{1}, \lambda_{ssim}, \lambda_{l}, \lambda_{s,b}, \lambda_{s,r}\}$ are set to $\{0.8, 0.2, 0.0001, 0.01, 0.01\}$, respectively.

\begin{table}[tb]
\centering
\caption{Quantitative results for novel view synthesis on NeRF synthetic dataset}
\label{tab:exp_nerf}
\scalebox{0.8}{
\begin{tabular}{@{}lccccc@{}}
\toprule[1.5pt]
 & Geometry & Relightable & PSNR $\uparrow$ & SSIM $\uparrow$ & LPIPS $\downarrow$ \\ 
\midrule
NPBG~\cite{aliev2020neural}         & point & \ding{56}         & 28.10      & 0.923     & 0.077  \\
NPBG++~\cite{rakhimov2022npbg++}    & point & \ding{56}         & 28.12      & 0.928     & 0.076  \\
FreqPCR~\cite{zhang2023frequency}   & point & \ding{56}         & \underline{31.24}      & \underline{0.950}     & \underline{0.049}  \\
3DGS~\cite{kerbl20233d}             & point & \ding{56}         & \textbf{33.88}      & \textbf{0.970}     & \textbf{0.031} \\
\midrule
PhySG~\cite{zhang2021physg}          & neural & \ding{52}        & 18.91      & 0.847     & 0.182  \\
NeILF++~\cite{zhang2023neilfpp}      & neural & \ding{52}        & 26.37      & 0.911     & 0.091      \\
Nvdiffrec~\cite{munkberg2022extracting} & mesh & \ding{52}        & \underline{29.05}      & \underline{0.939}     & \underline{0.081}  \\
R3DG(Ours)                                 & point & \ding{52}        & \textbf{31.22}      & \textbf{0.959}     & \textbf{0.039}  \\
\bottomrule[1.5pt]
\end{tabular}
}
\end{table}
\begin{figure}[tb]
\centering
\resizebox{\linewidth}{!}{
\begin{subfigure}[b]{0.32\linewidth}
    \includegraphics[width=\linewidth]{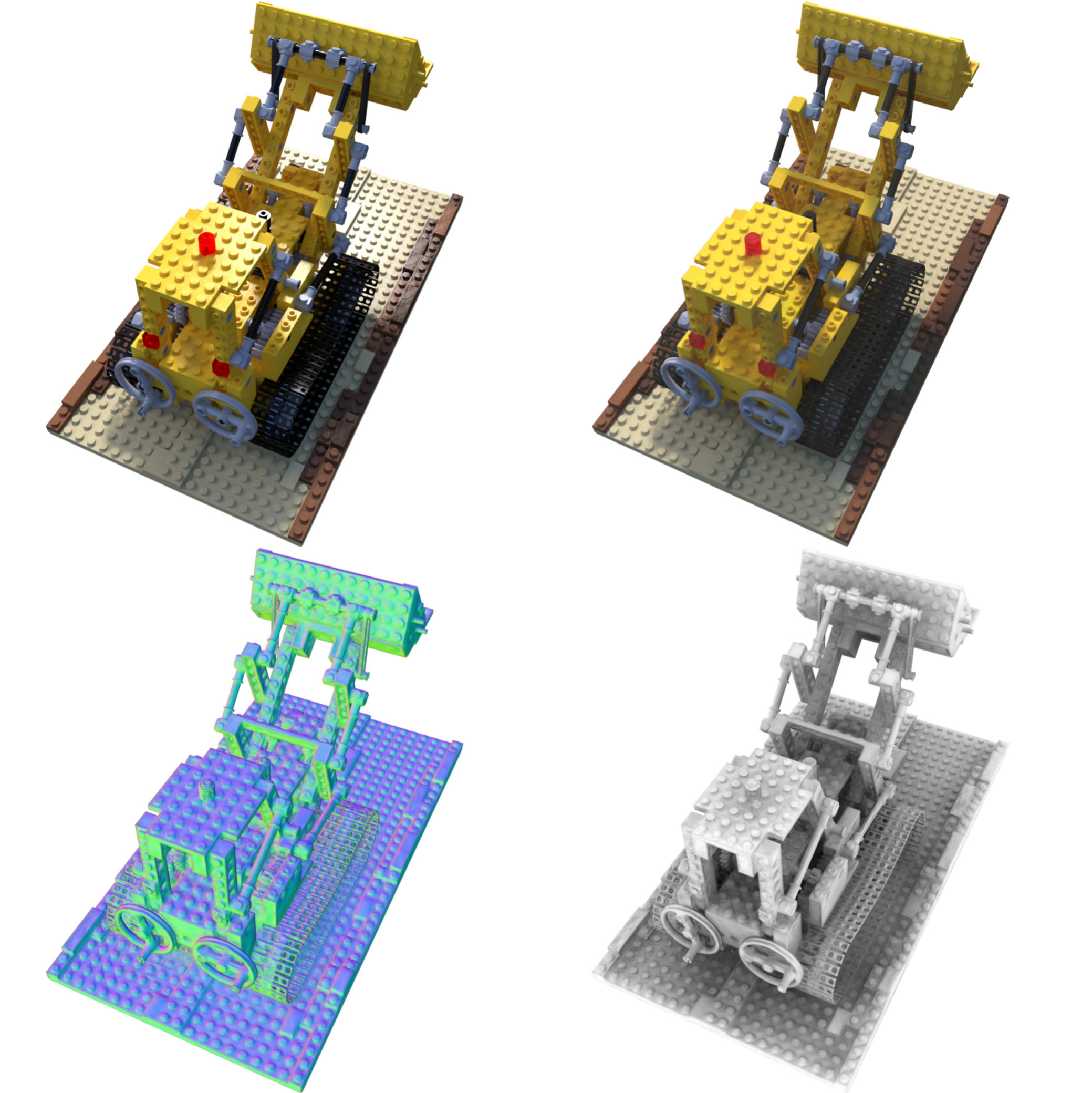}
    \caption{Lego}
\end{subfigure}
\begin{subfigure}[b]{0.32\linewidth}
    \includegraphics[width=\linewidth]{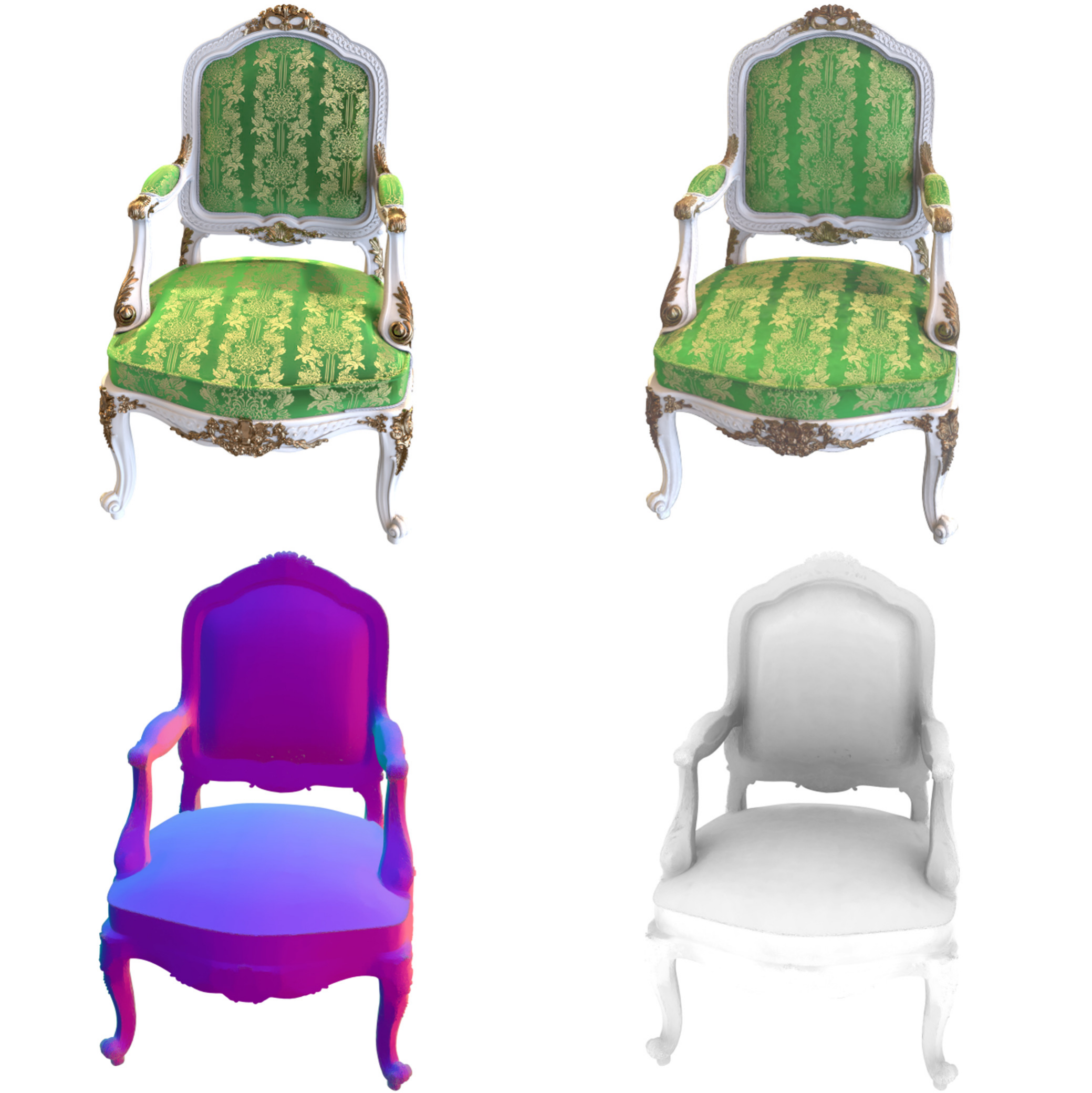}
    \caption{Chair}
\end{subfigure}
\begin{subfigure}[b]{0.32\linewidth}
    \includegraphics[width=\linewidth]{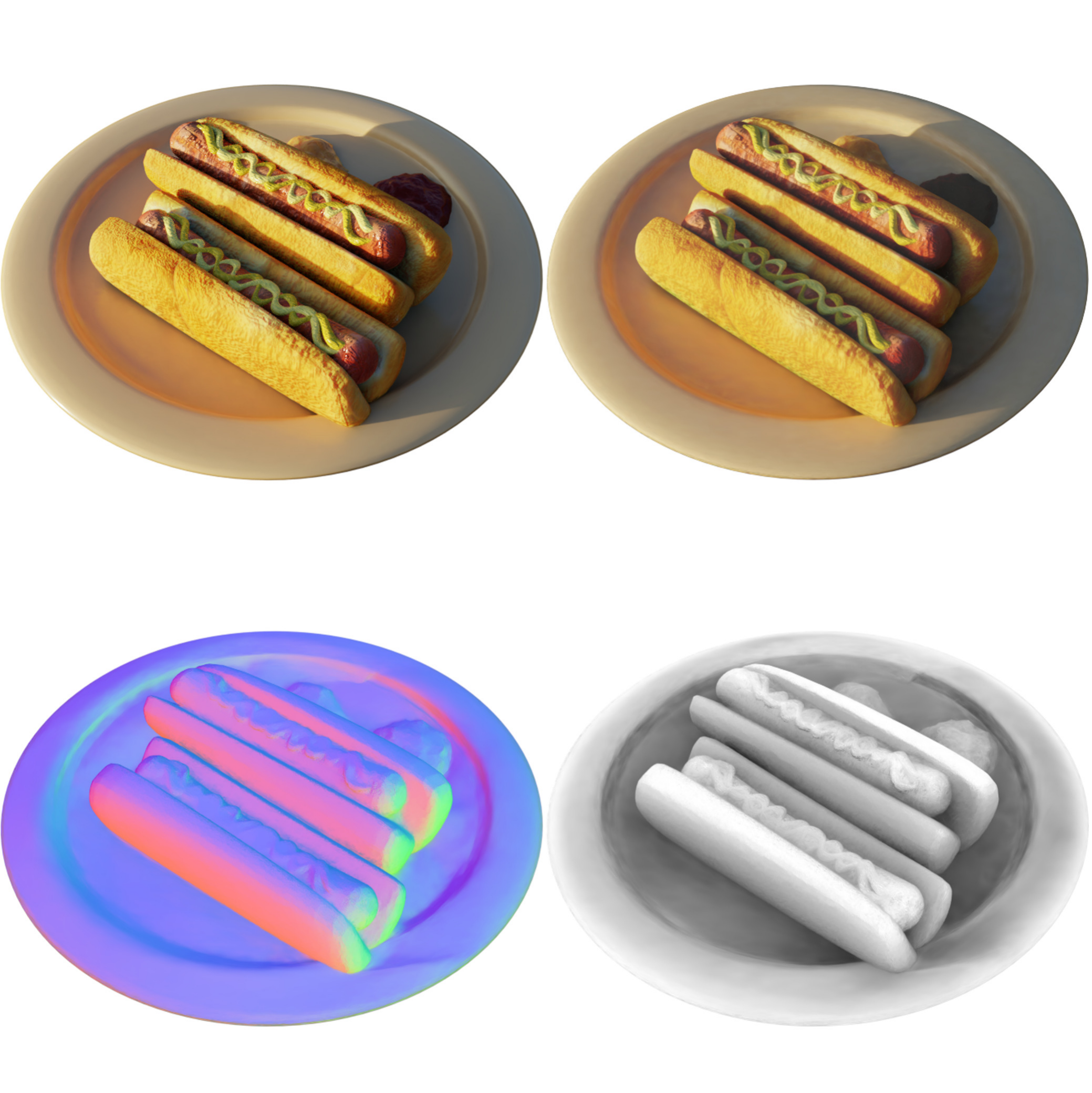}
    \caption{Hotdog}
\end{subfigure}
}
\caption{\textbf{Visualizations on NeRF synthetic dataset}~\cite{mildenhall2020nerf}. Each scene is displayed in an order from left to right and from top to bottom: Ground Truth, PBR Image, Normal Map and Ambient Occlusion Map.}
\label{fig:nerf_synthetic}
\end{figure}

\subsection{Performance on Novel View Synthesis}
\label{para:exp_synthetic}
We first evaluate the novel view synthesis (NVS) of the physically based rendering (PBR) on NeRF synthetic dataset~\cite{mildenhall2020nerf}. We compare both \textit{point-based} rendering methods and \textit{relightable} rendering approaches. We report average metrics across all scenes in Tab.~\ref{tab:exp_nerf}, including peak signal-to-noise ratio (PSNR), structural similarity index measure (SSIM), and learned perceptual image patch similarity (LPIPS). Compared with existing \textit{point-based} rendering methods, our R3DG demonstrates superiority over most of them ~\cite{aliev2020neural,rakhimov2022npbg++,zhang2023frequency}. While our performance slightly marginally trails that of vanilla 3DGS~\cite{kerbl20233d}, it is noteworthy that our approach excels in relighting capability, presenting a significant advantage. Furthermore, in comparison to other \textit{relightable} methods~\cite{zhang2021physg,munkberg2022extracting,zhang2023neilfpp}, our method showcases significantly better NVS quality. 

In Fig.~\ref{fig:nerf_synthetic}, we visualize the our reconstruction results in NeRF synthetic dataset, including the PBR image, the normal map, and the pre-computed visibility illustrated in the form of the Ambient Occlusion (AO) map. As depicted in Fig.~\ref{fig:nerf_synthetic}, our approach successfully achieves realistic PBR rendering, smooth normal estimation, and accurate visibility solving on discrete point clouds.

\begin{table}[t]
\centering
\caption{Quantitative results on Synthetic4Relight dataset~\cite{zhang2022modeling}}
\label{tab:exp_syn4}
\resizebox{\linewidth}{!}{
\begin{tabular}{lcccccccccccc}
\toprule[1.5px]
     & \multicolumn{3}{c}{\textbf{Novel View Synthesis}} & \multicolumn{3}{c}{\textbf{Relighting}} & \multicolumn{3}{c}{\textbf{Albedo}} & \textbf{Roughness} & Time \\
                  & PSNR↑  & SSIM↑ & LPIPS↓ & PSNR↑    & SSIM↑    & LPIPS↓   & PSNR↑   & SSIM↑  & LPIPS↓  & MSE↓  & (hours)    \\ \hline
NerFactor~\cite{zhang2021nerfactor}         & 22.80  & 0.916 & 0.150  & 21.54    & 0.875    & 0.171    & 19.49   & 0.864  & 0.206   & N/A   & >48    \\
Nvdifferc-MC~\cite{hasselgren2022shape}      & 34.29  & 0.967 & 0.068  & 24.22    & 0.943    & 0.078    & 29.61   & 0.945  & 0.075   & 0.009  & 4.17   \\
InvRender~\cite{zhang2022modeling}         & 30.74  & 0.953 & 0.086  & 28.67    & 0.950    & 0.091    & 28.28   & 0.935  & 0.072   & \textbf{0.008} & 14.3     \\
TensorIR~\cite{jin2023tensoir}          & 35.80  & 0.978 & 0.049  & 29.69    & 0.951    & 0.079    & \textbf{30.58}   & 0.946  & 0.065   & 0.015  & 3.24   \\
R3DG (Ours)    &   \textbf{36.80}  &  \textbf{0.982}   &   \textbf{0.028}     &  \textbf{31.00}       &    \textbf{0.964}      &    \textbf{0.050} & 28.31   &    \textbf{0.951}    &    \textbf{0.058}     &    0.013   &  \textbf{0.90}    \\ 
\bottomrule[1.5px]
\end{tabular}
}
\end{table}
\begin{figure}[t]
\centering

\resizebox{\linewidth}{!}{
\begin{tabular}{ccccccc}

& Rendering & Albedo & Roughness & Env.map & Relighting1 & Relighting2  \\
\raisebox{0.07\textwidth}[0pt][0pt]{\rotatebox[origin=c]{90}{\footnotesize InvRender}} & 
\includegraphics[width=0.16\textwidth]{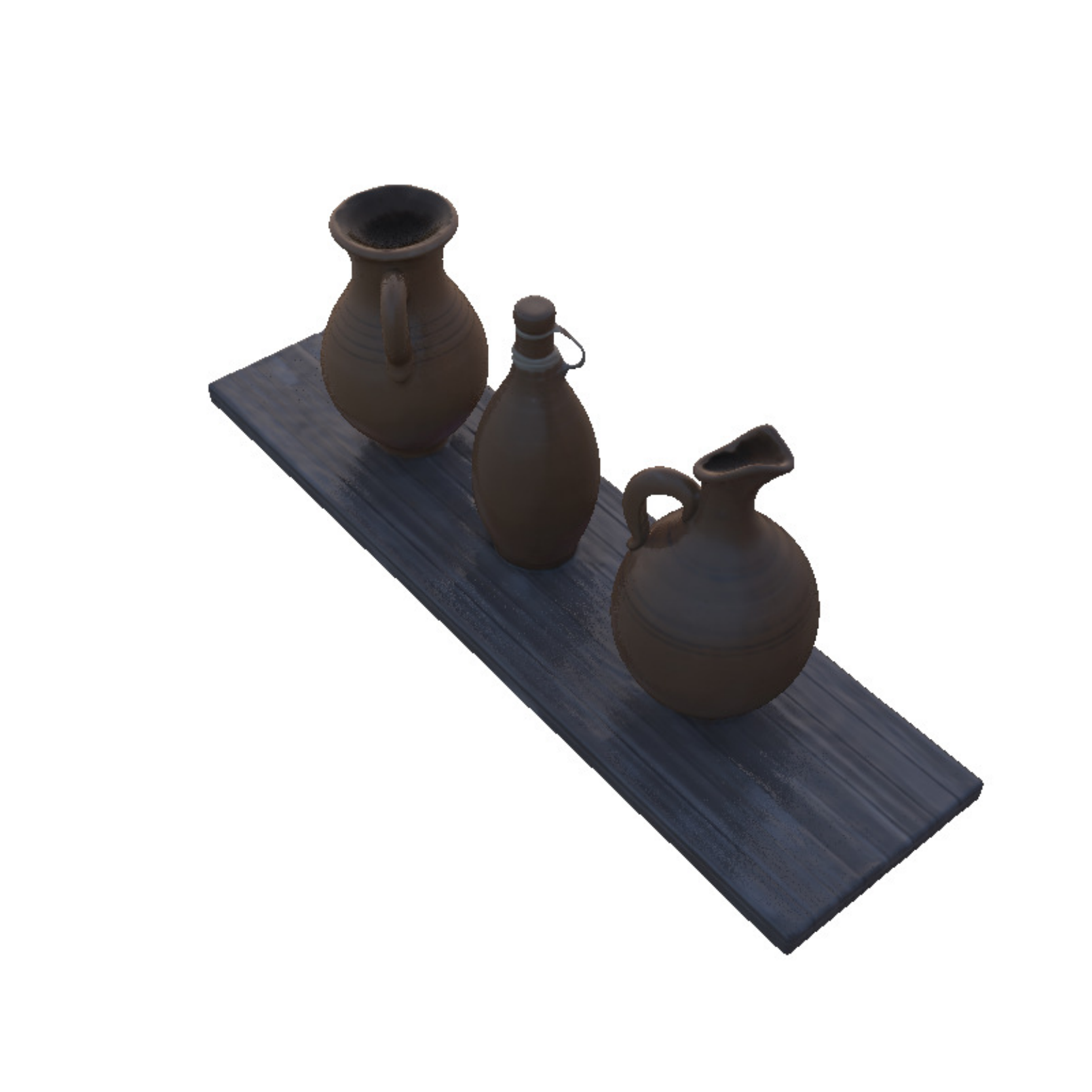} & \includegraphics[width=0.16\textwidth]{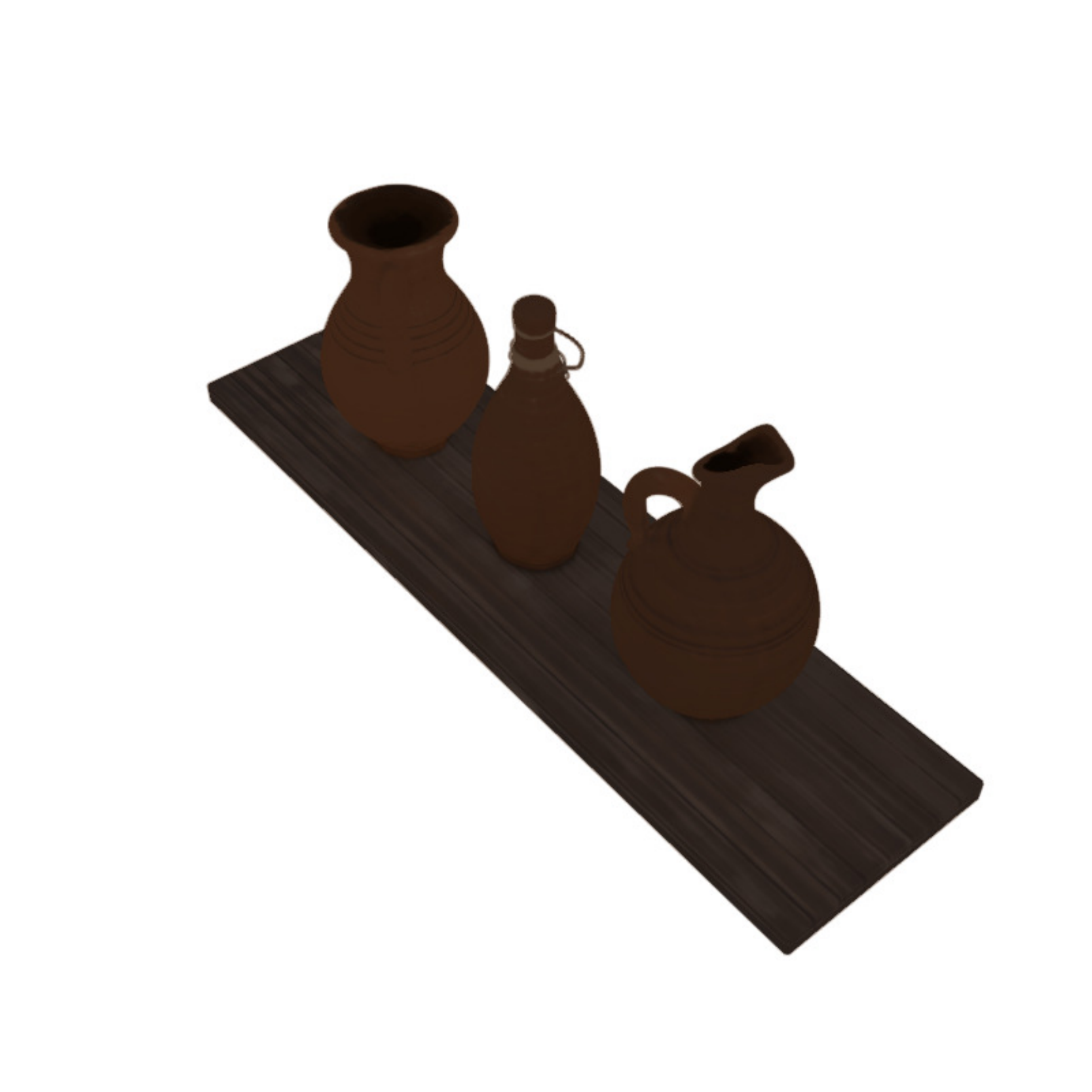} & 
\includegraphics[width=0.16\textwidth]{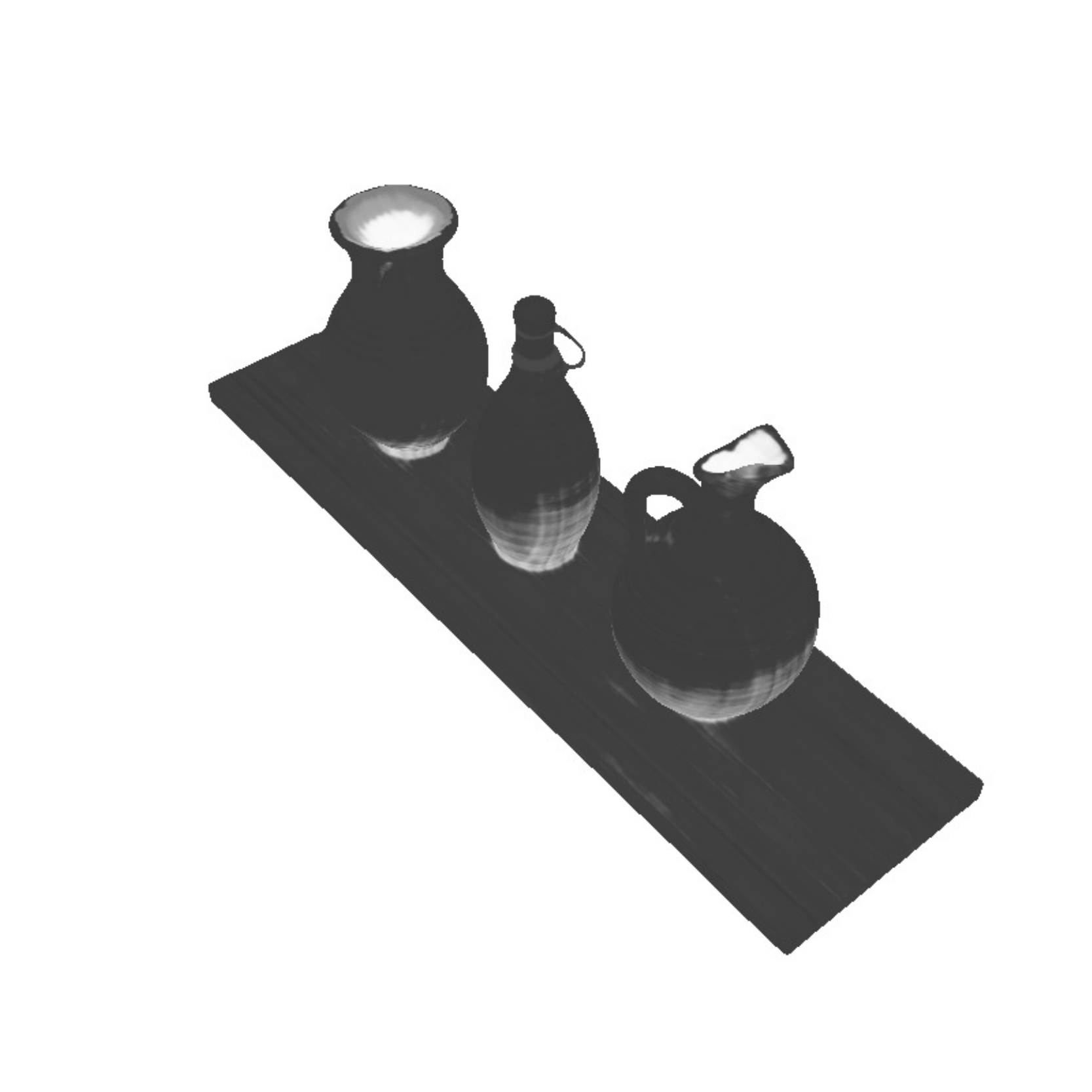} & 
\includegraphics[width=0.16\textwidth]{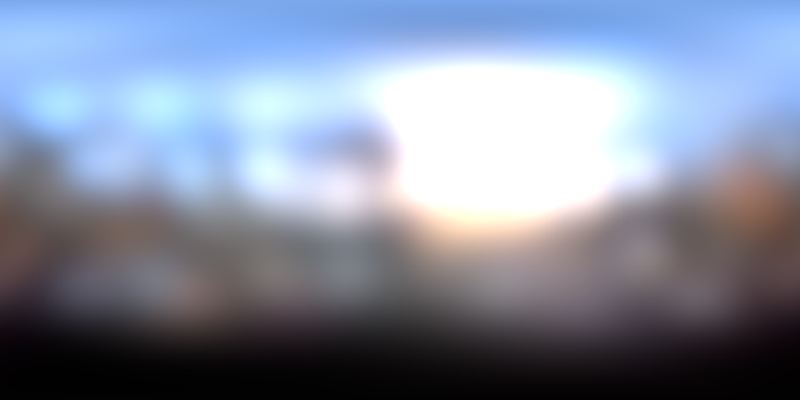} &
\includegraphics[width=0.16\textwidth]{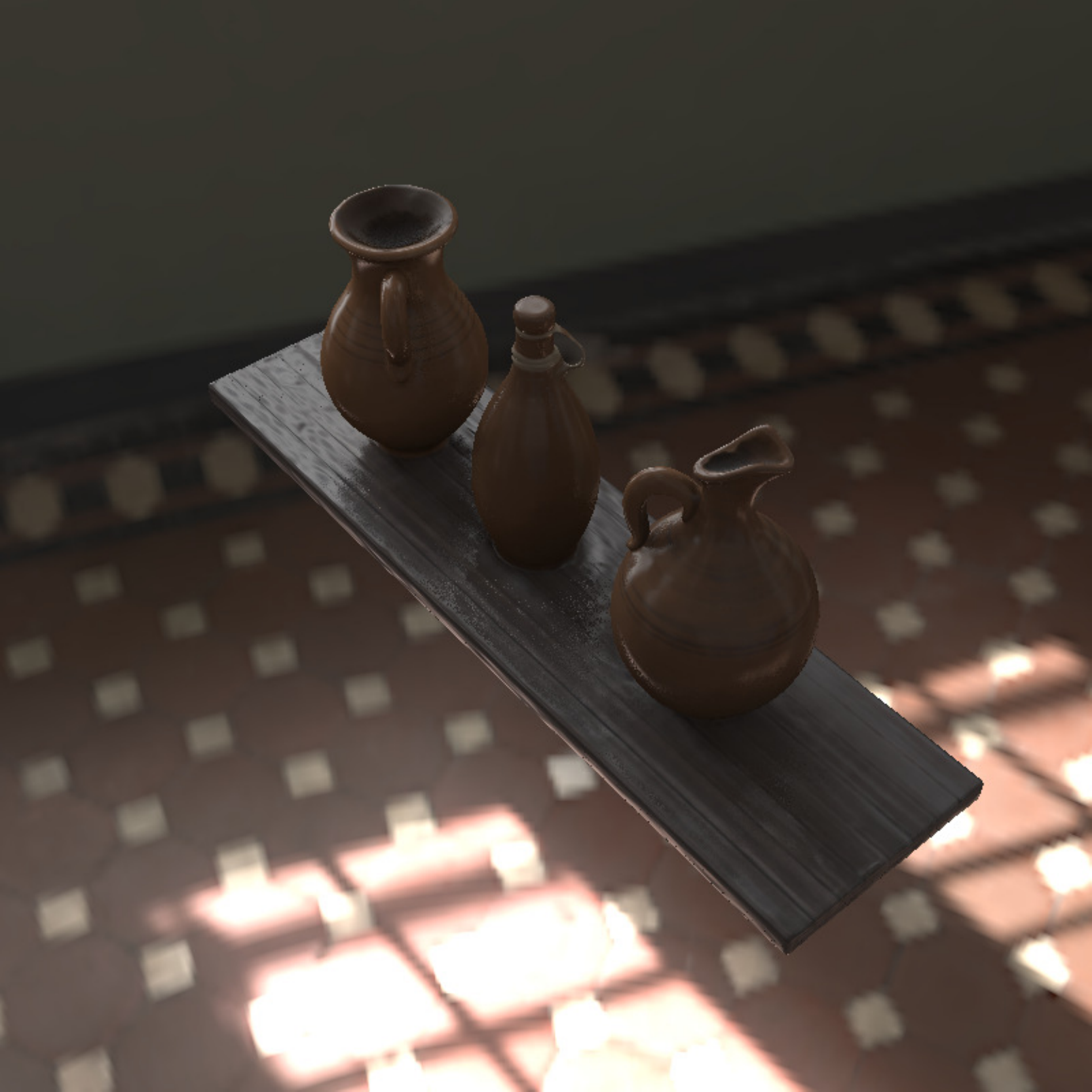} & 
\includegraphics[width=0.16\textwidth]{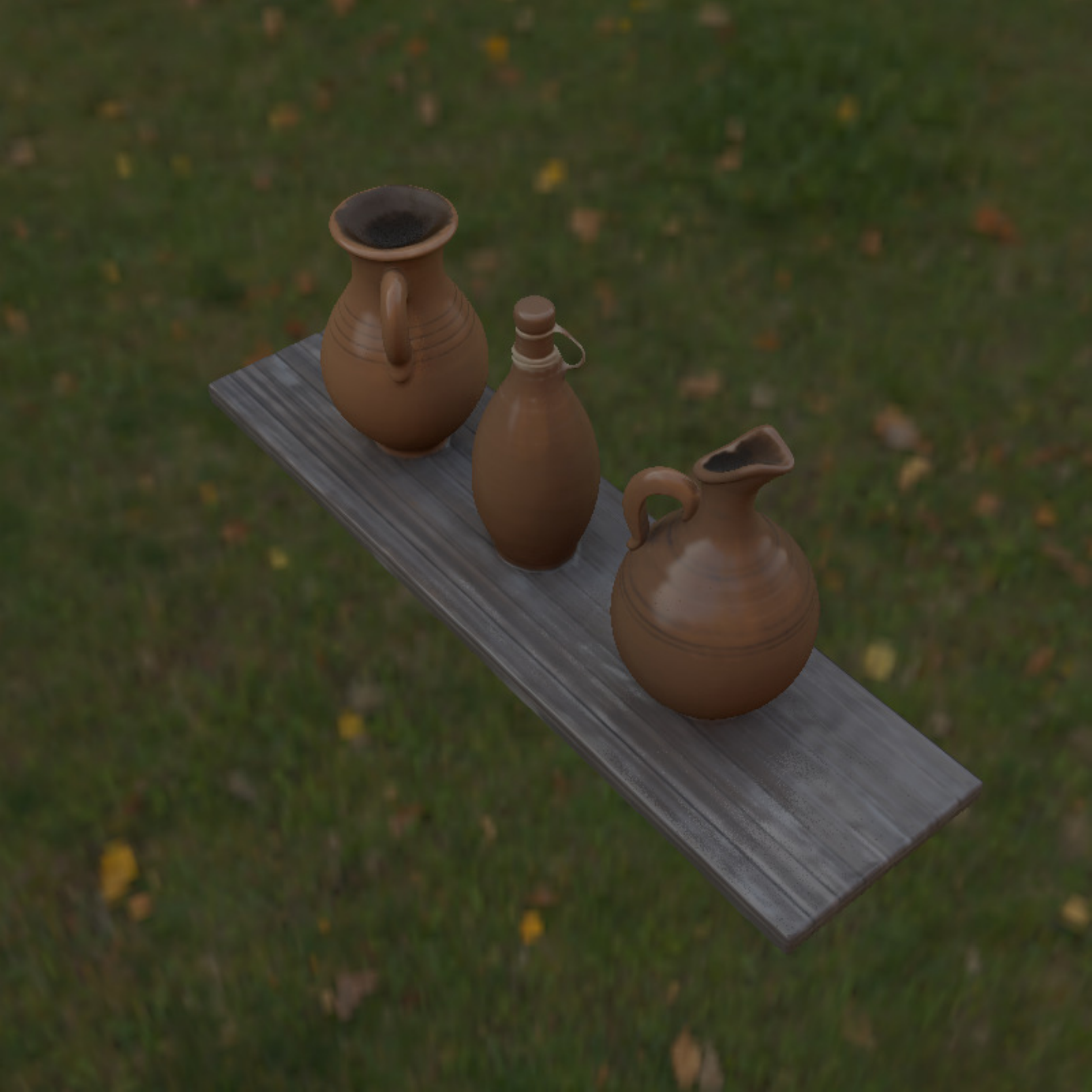} 
\\

\raisebox{0.07\textwidth}[0pt][0pt]{\rotatebox[origin=c]{90}{\footnotesize Nvdifferc-MC}} & 
\includegraphics[width=0.16\textwidth]{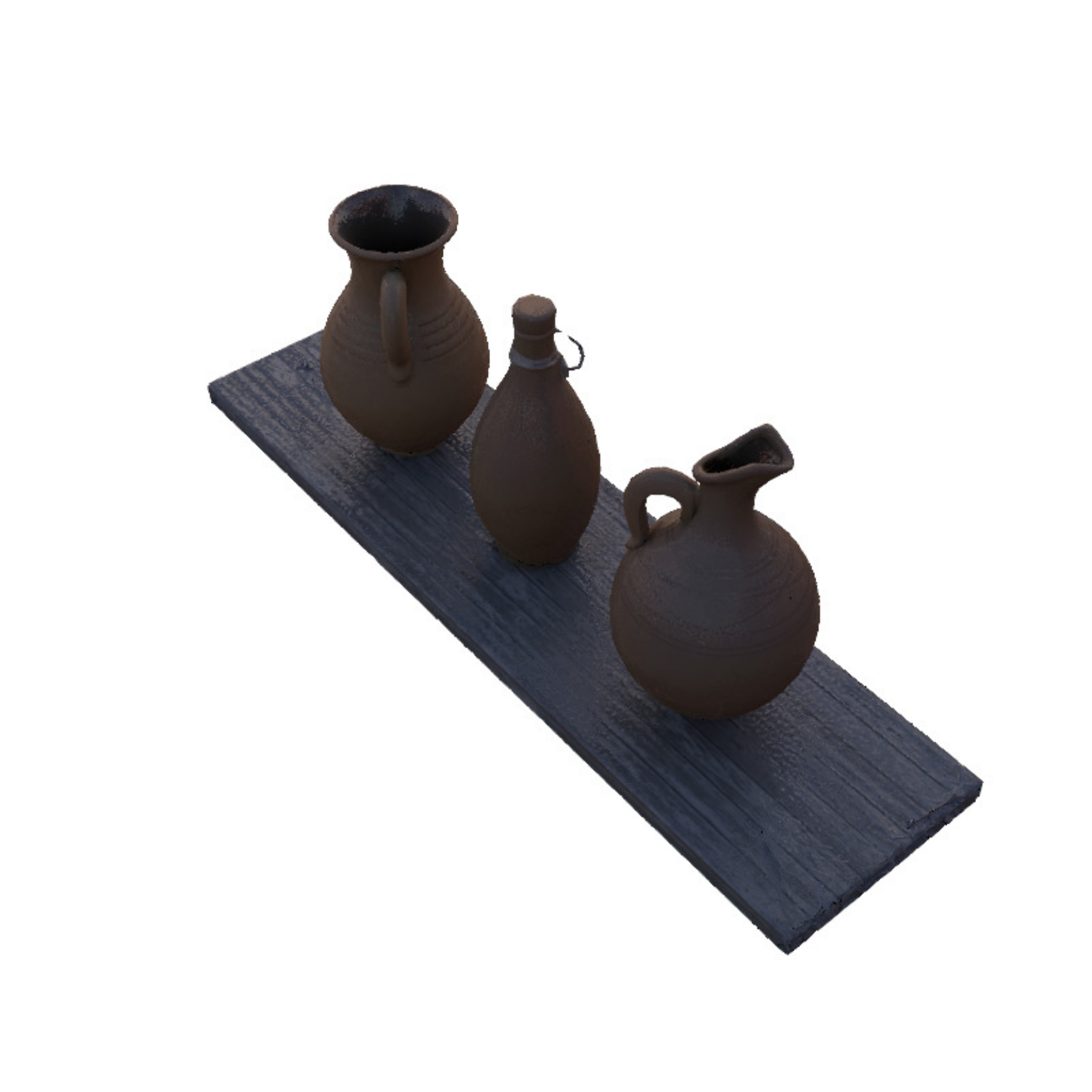} & \includegraphics[width=0.16\textwidth]{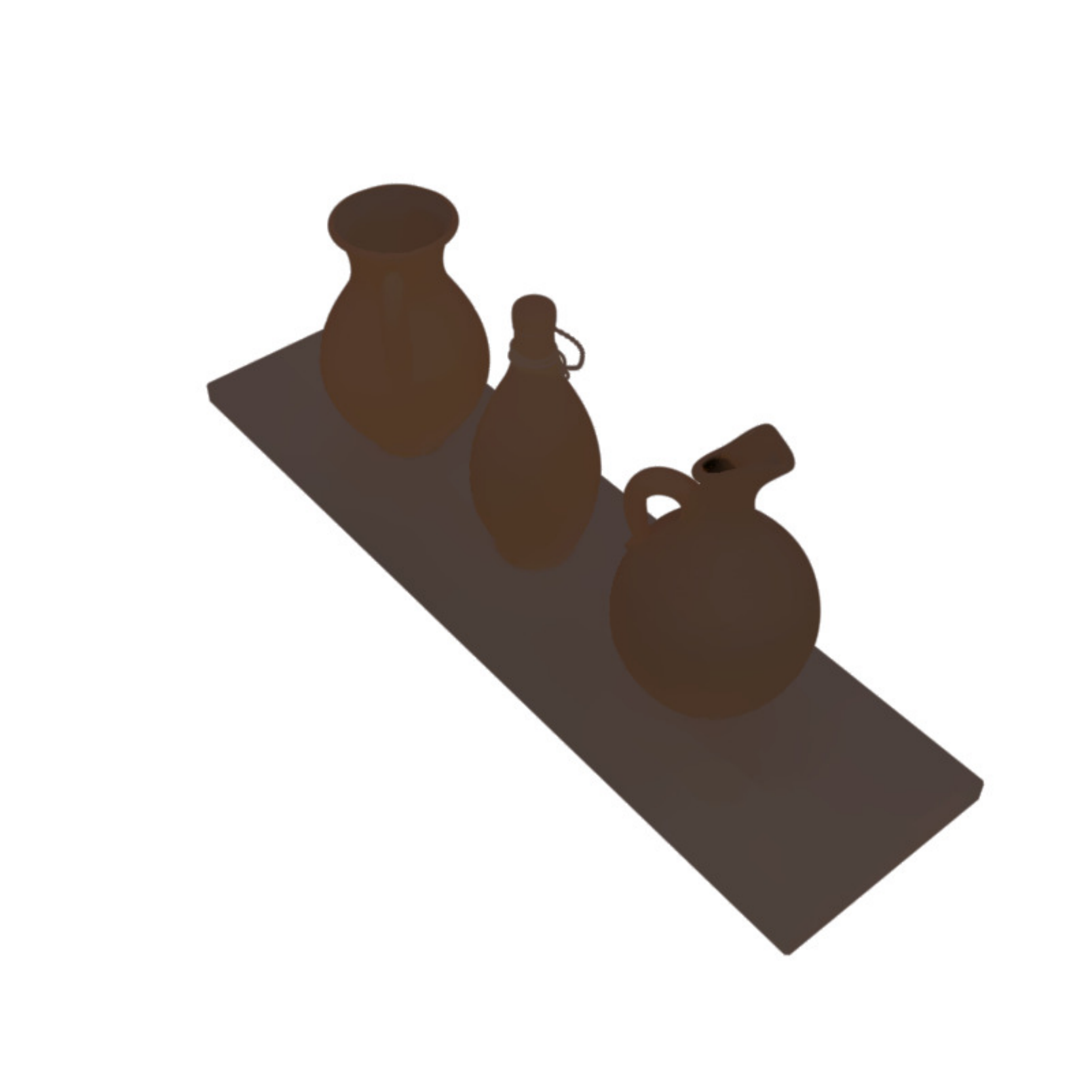} & 
\includegraphics[width=0.16\textwidth]{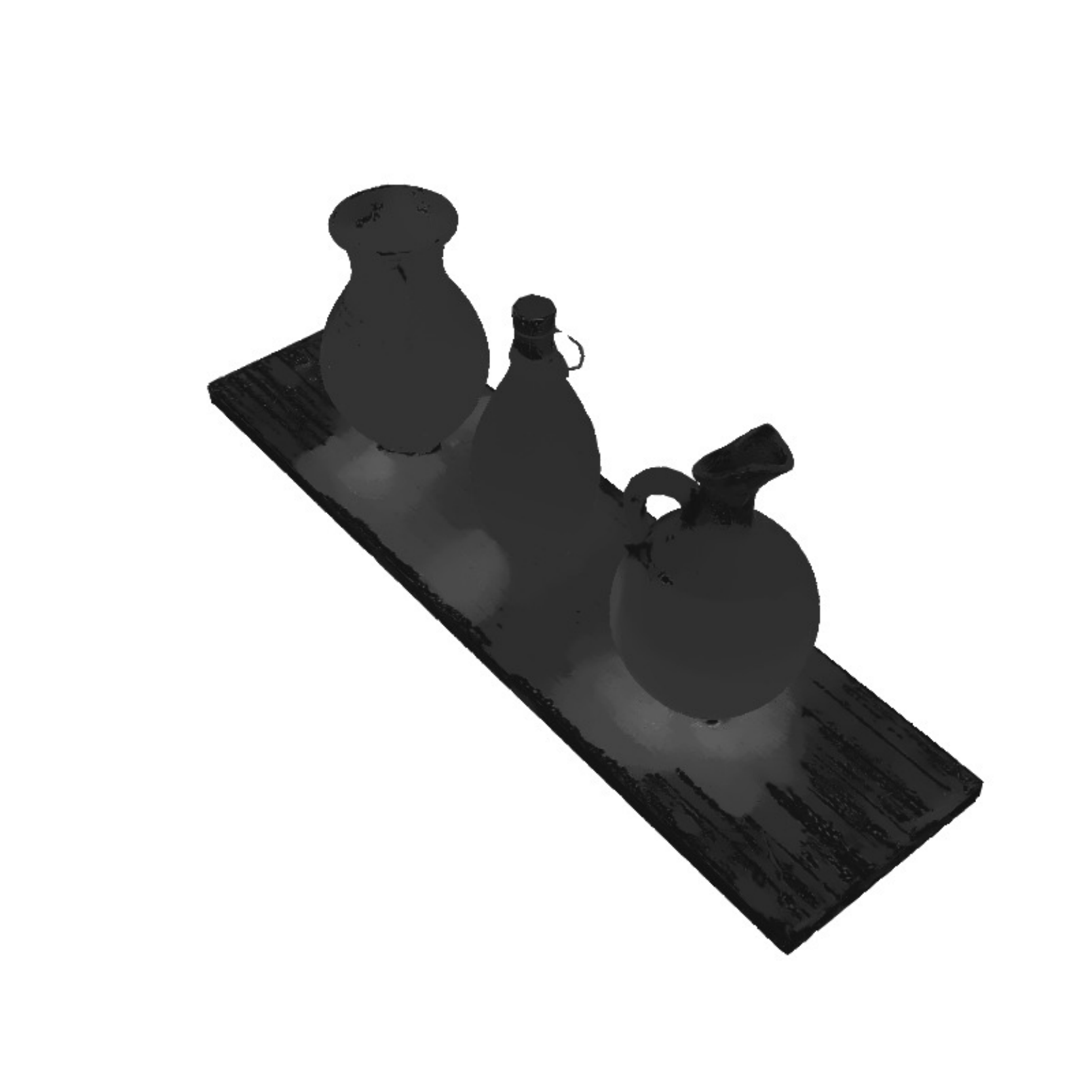} & 
\includegraphics[width=0.16\textwidth]{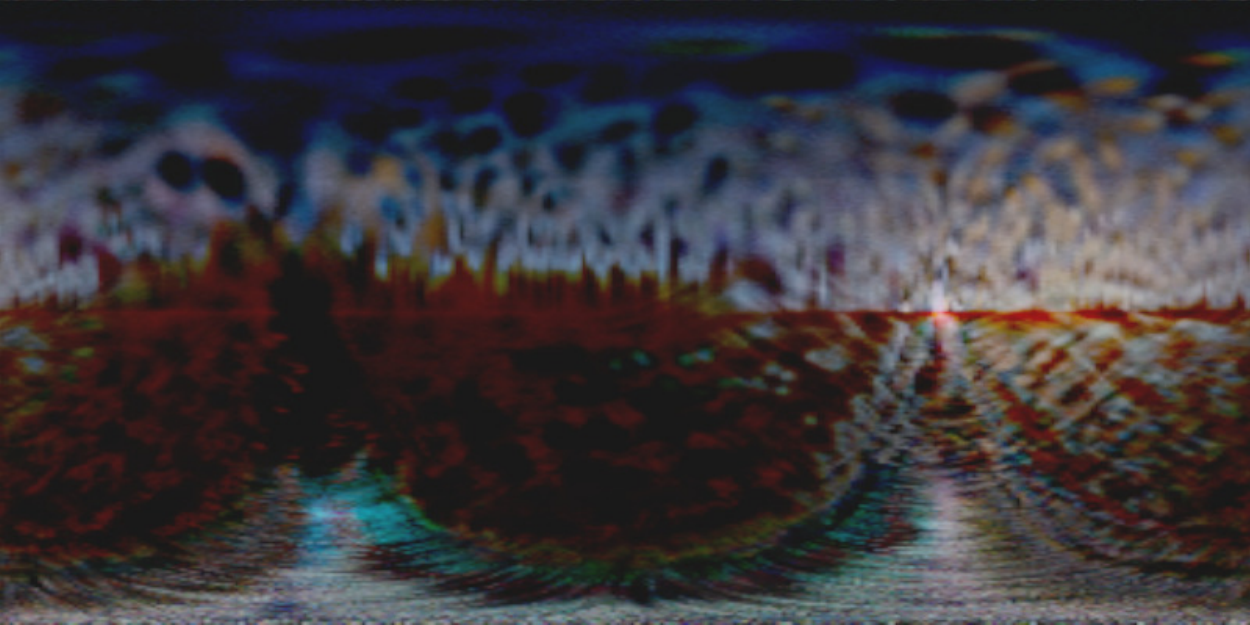} &
\includegraphics[width=0.16\textwidth]{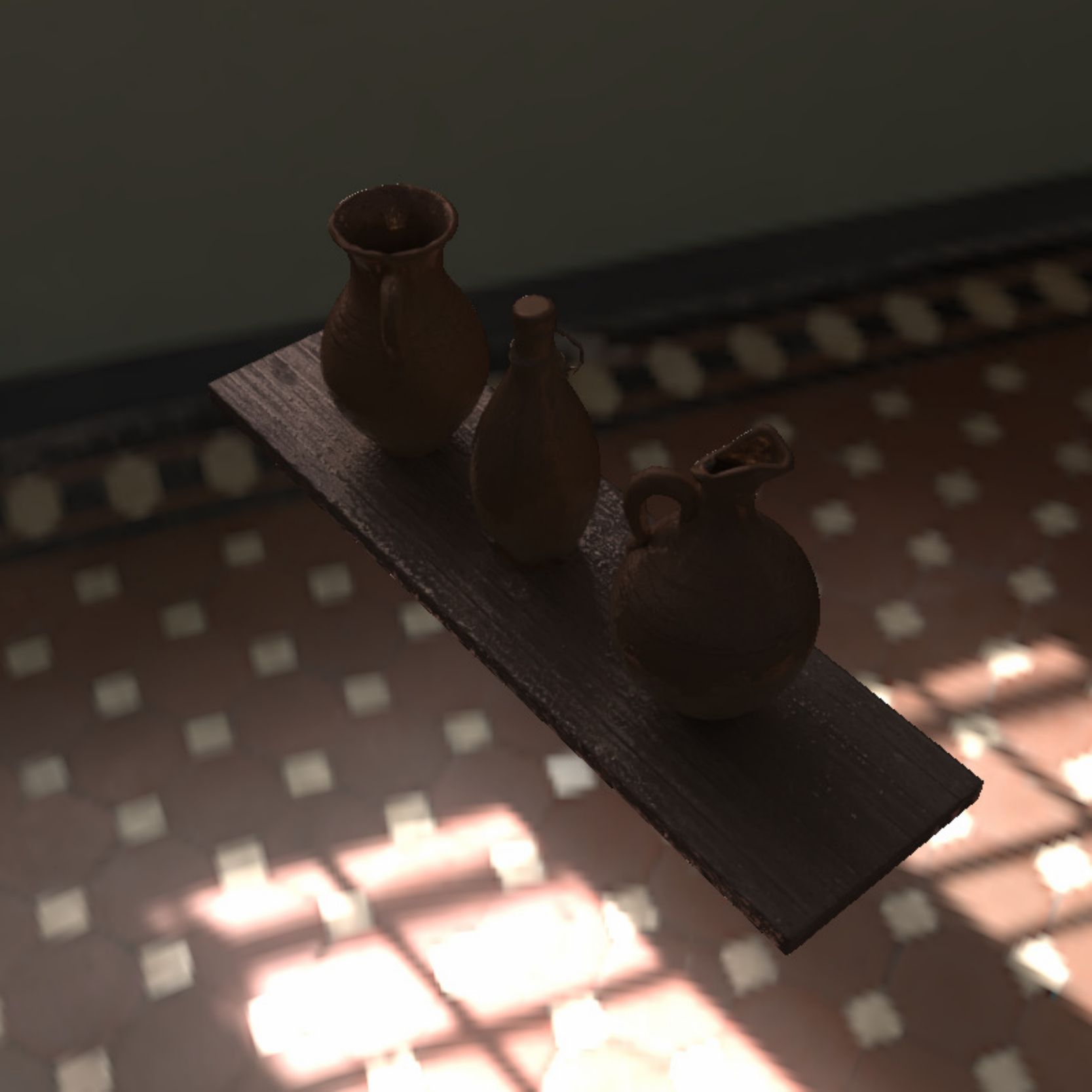} & 
\includegraphics[width=0.16\textwidth]{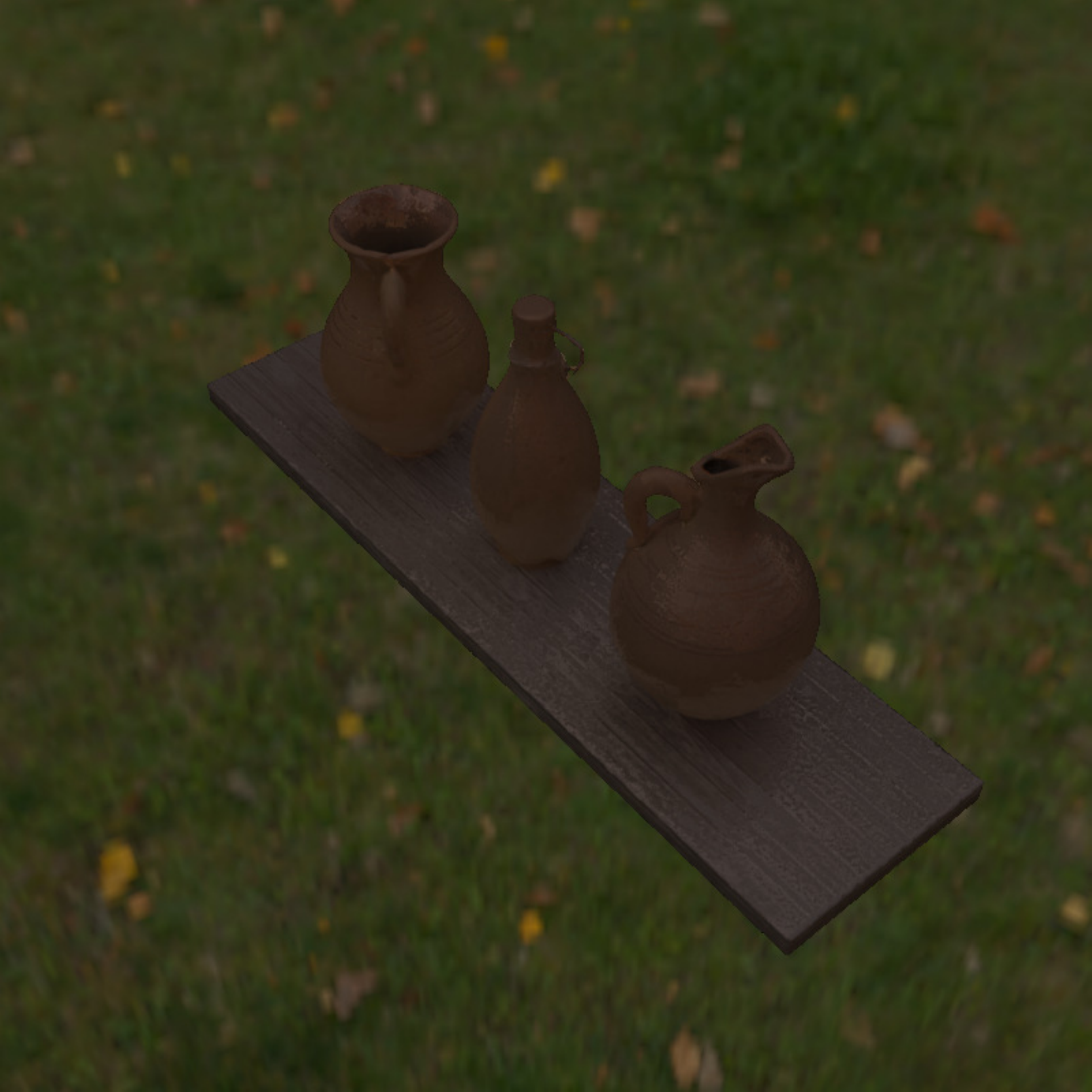} 
\\

\raisebox{0.07\textwidth}[0pt][0pt]{\rotatebox[origin=c]{90}{\footnotesize Ours}} & 
\includegraphics[width=0.16\textwidth]{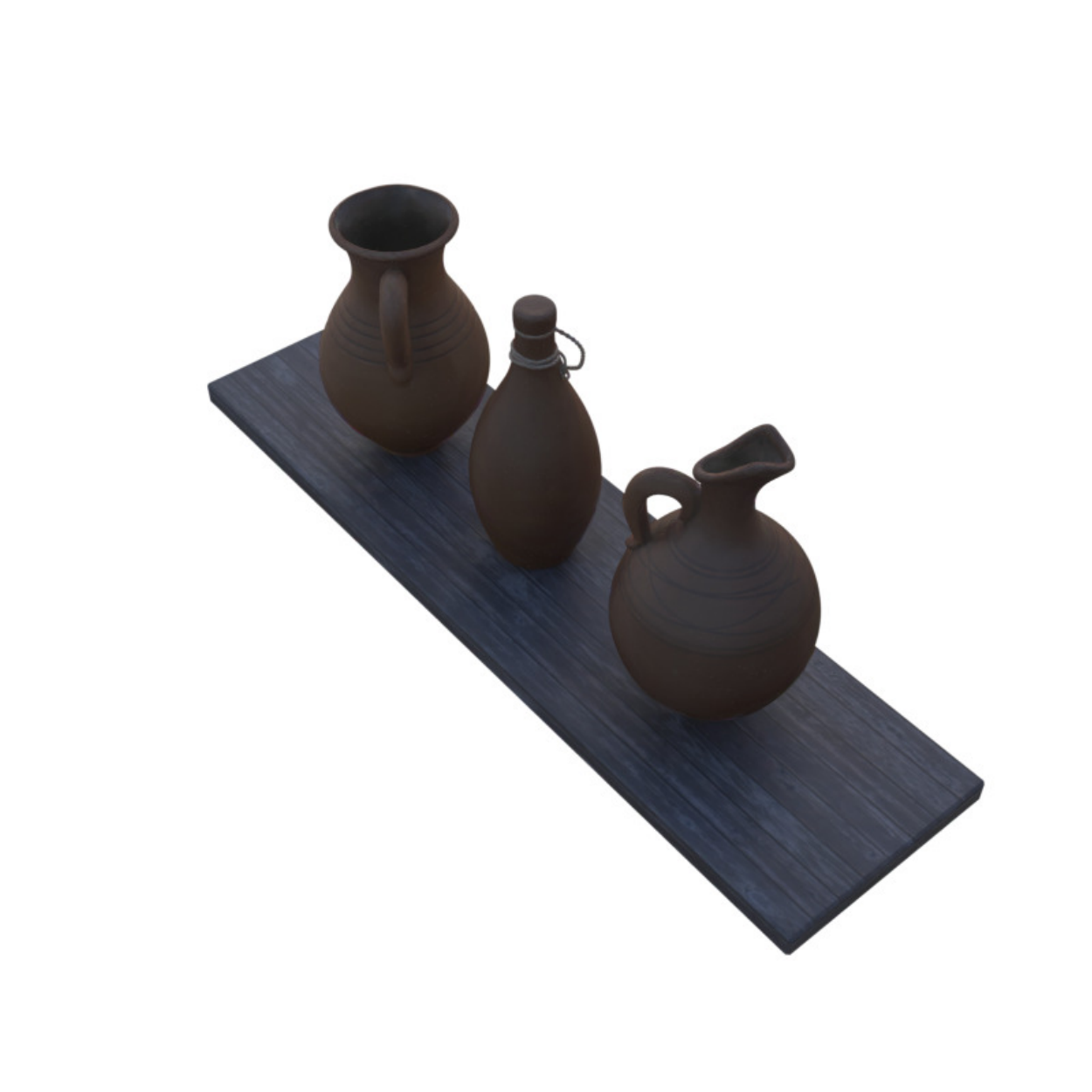} & \includegraphics[width=0.16\textwidth]{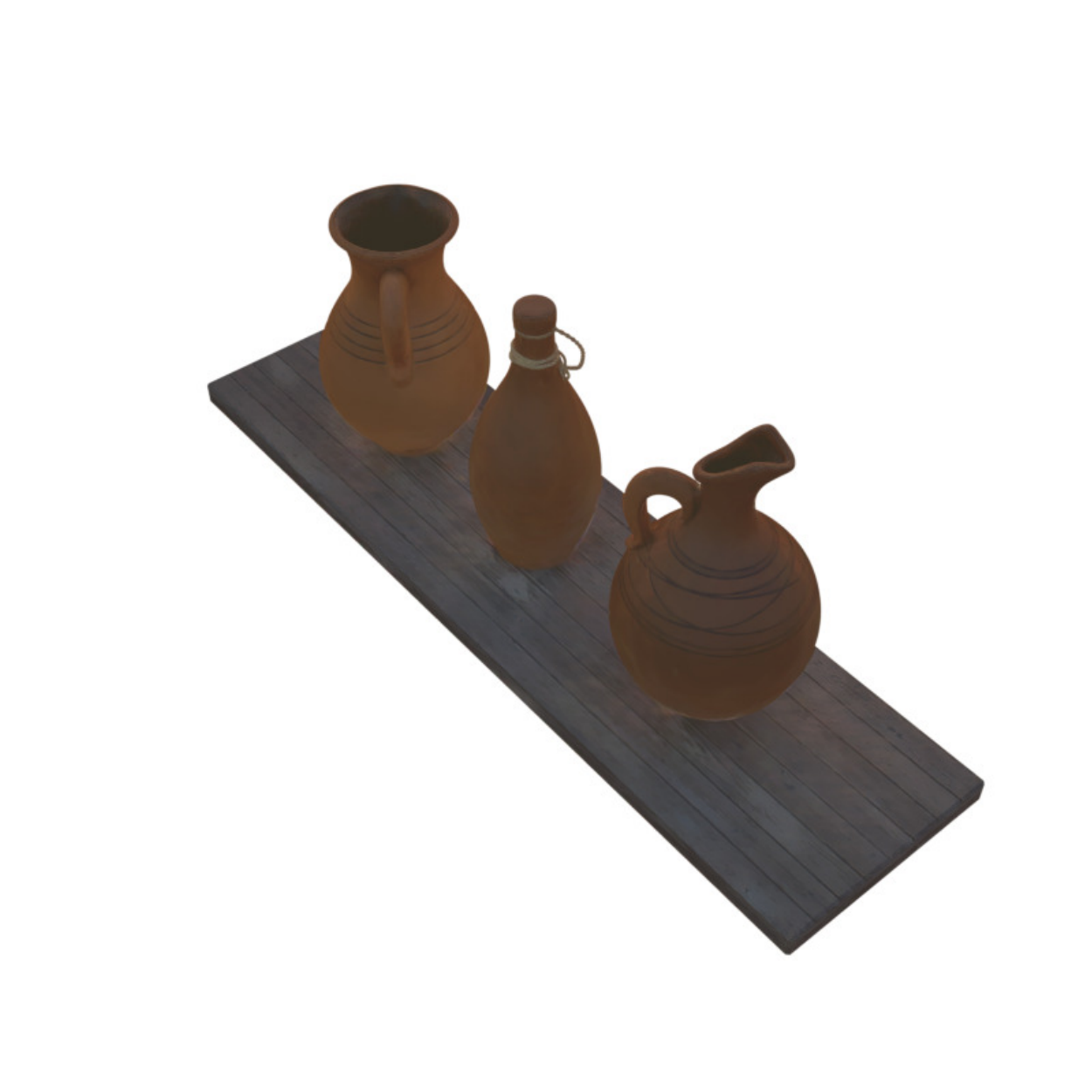} & 
\includegraphics[width=0.16\textwidth]{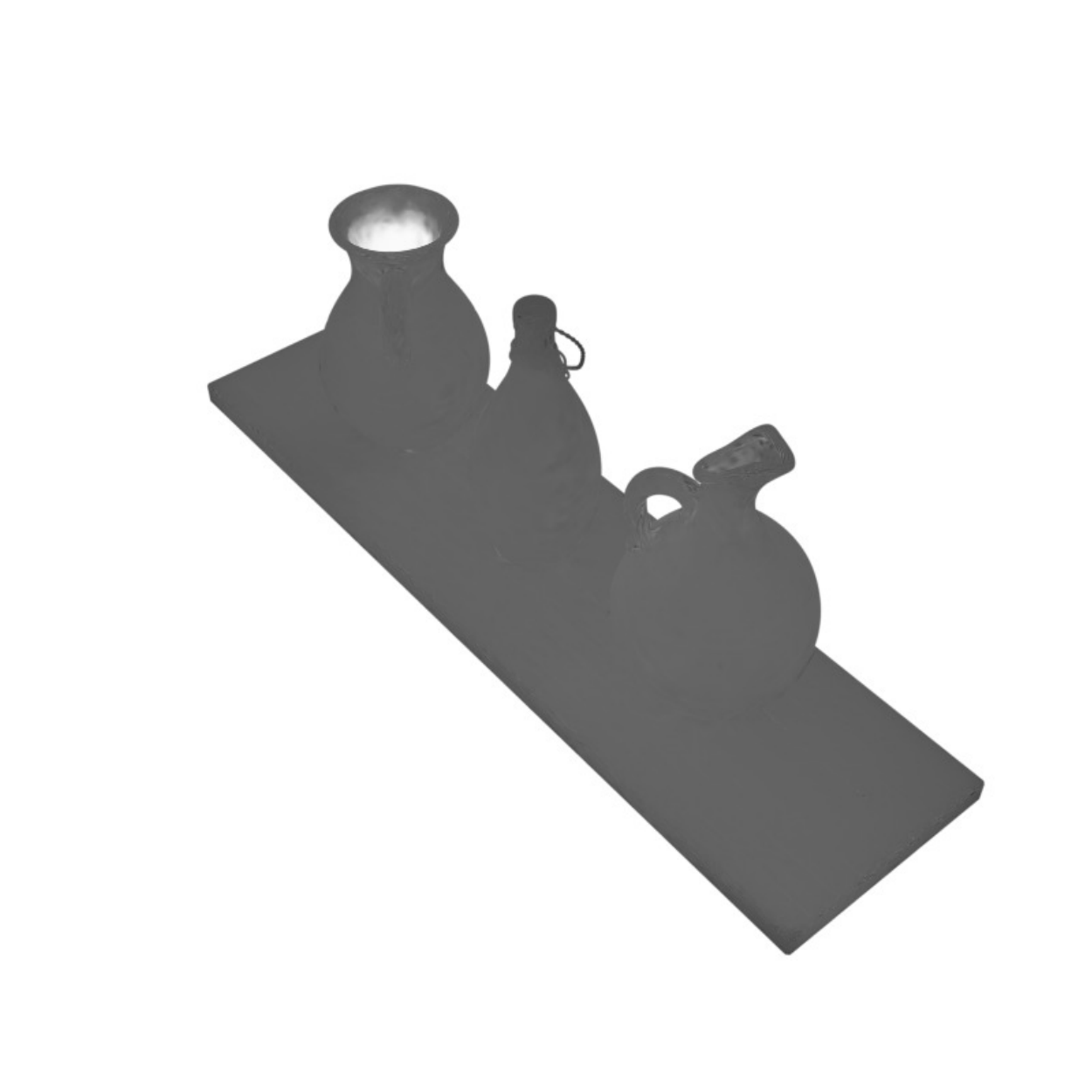} & 
\includegraphics[width=0.16\textwidth]{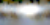} &
\includegraphics[width=0.16\textwidth]{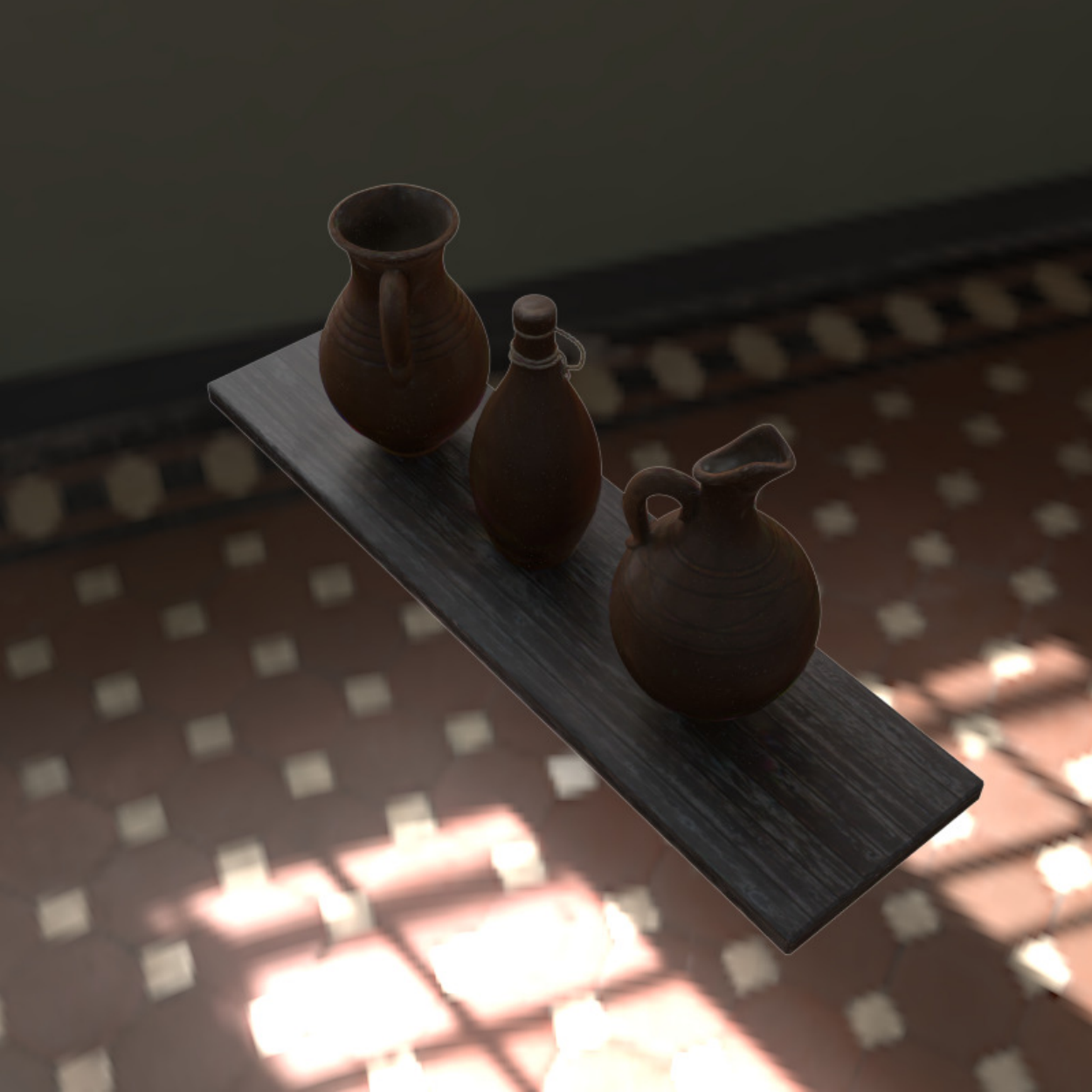} & 
\includegraphics[width=0.16\textwidth]{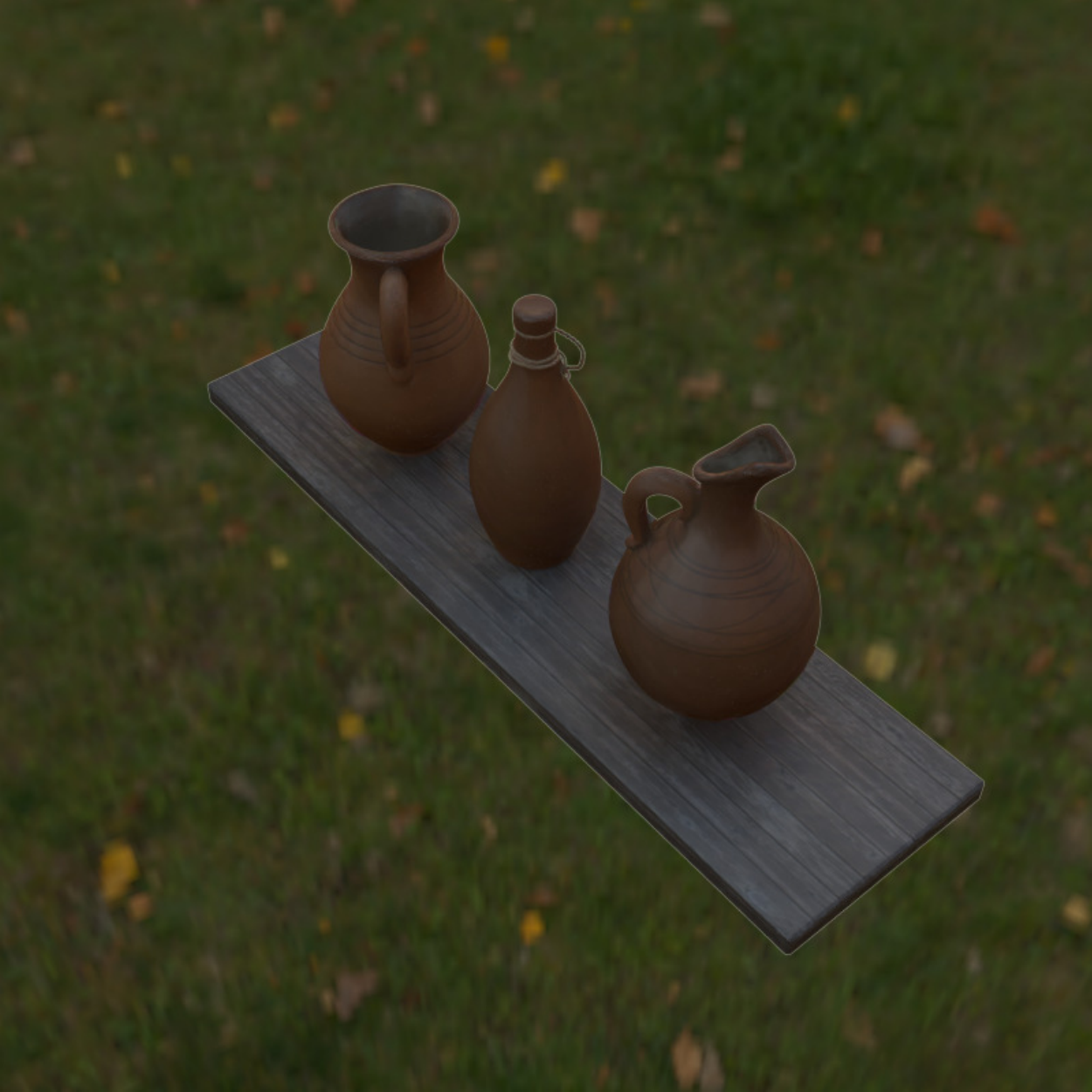} 
\\

\raisebox{0.07\textwidth}[0pt][0pt]{\rotatebox[origin=c]{90}{\footnotesize GT}} & 
\includegraphics[width=0.16\textwidth]{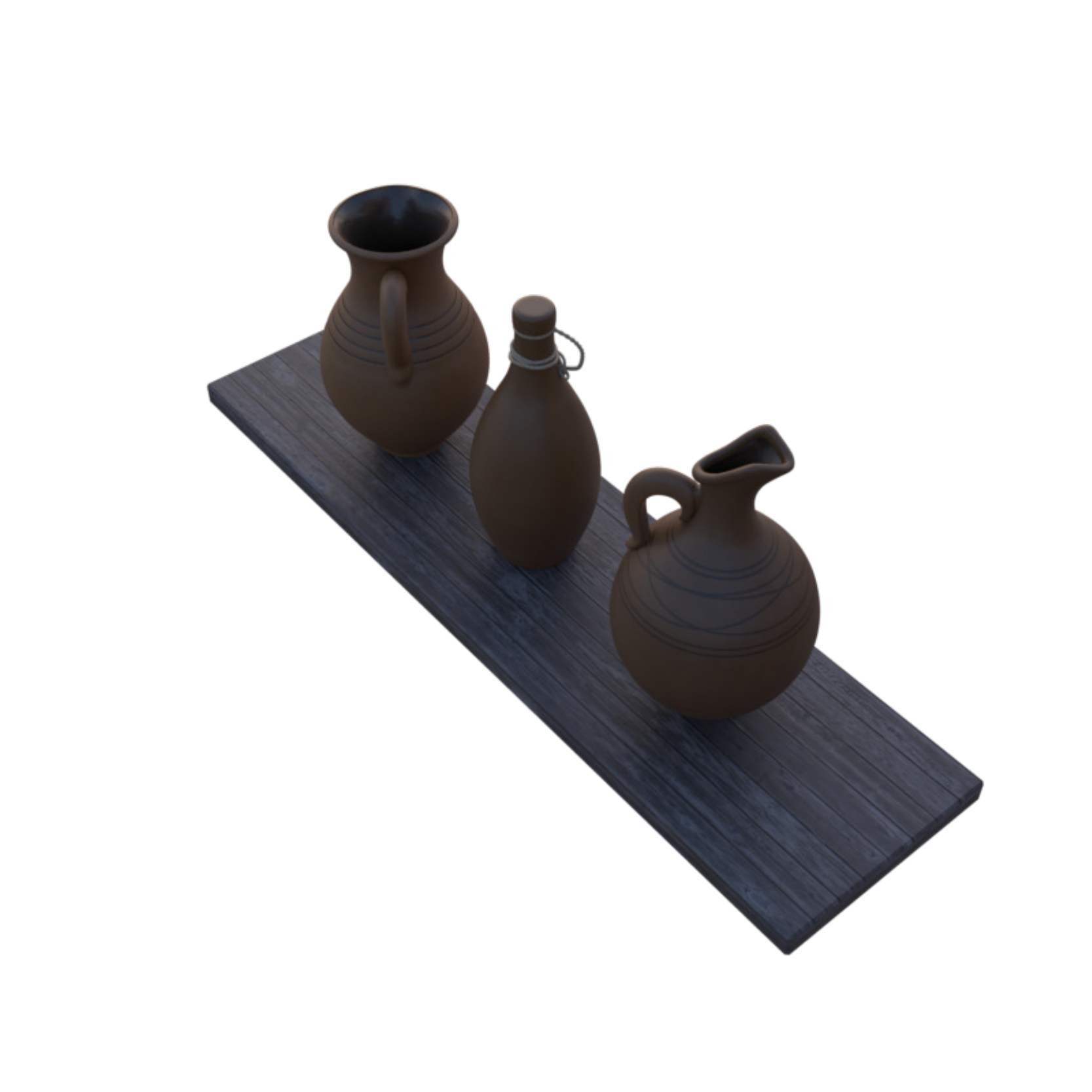} & \includegraphics[width=0.16\textwidth]{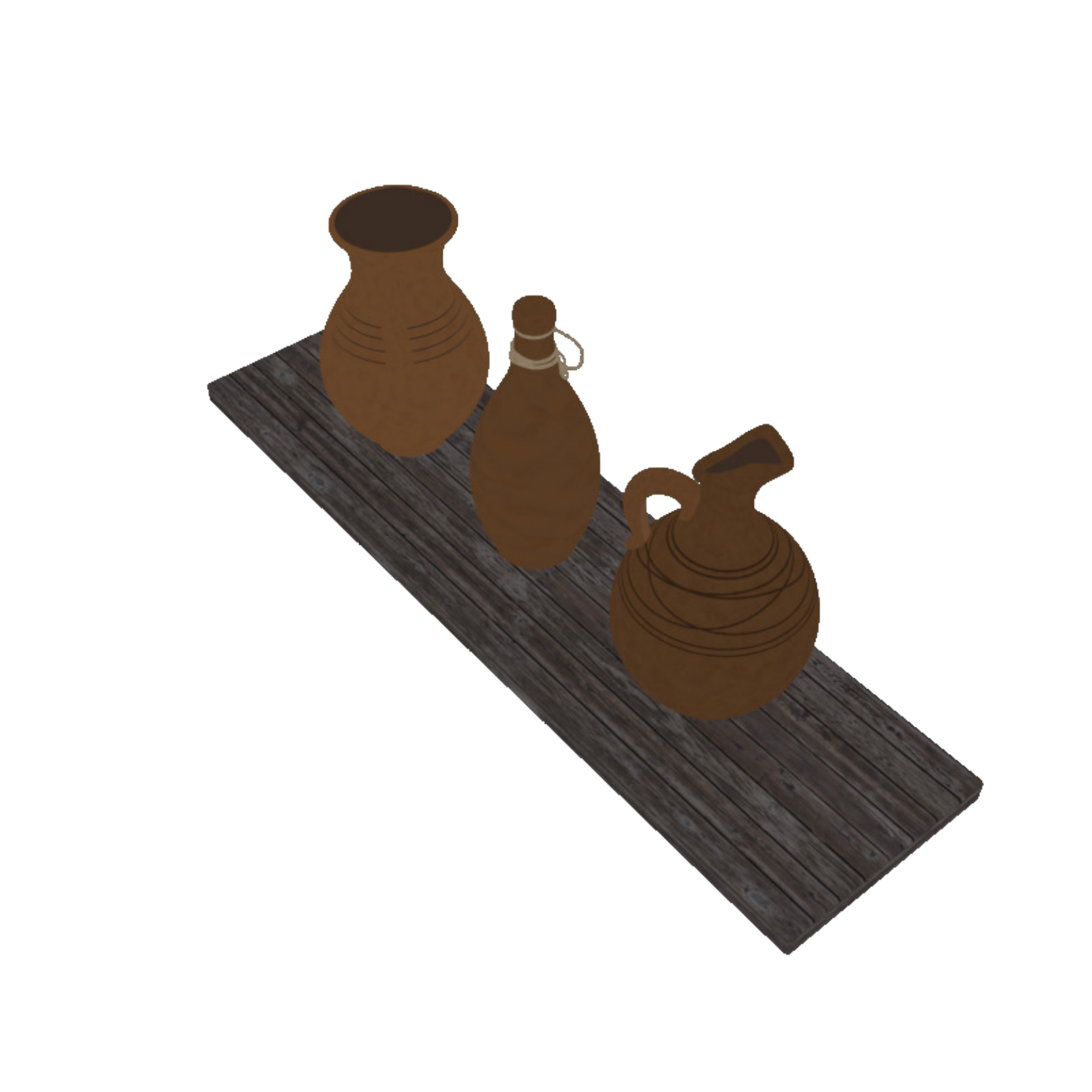} & 
\includegraphics[width=0.16\textwidth]{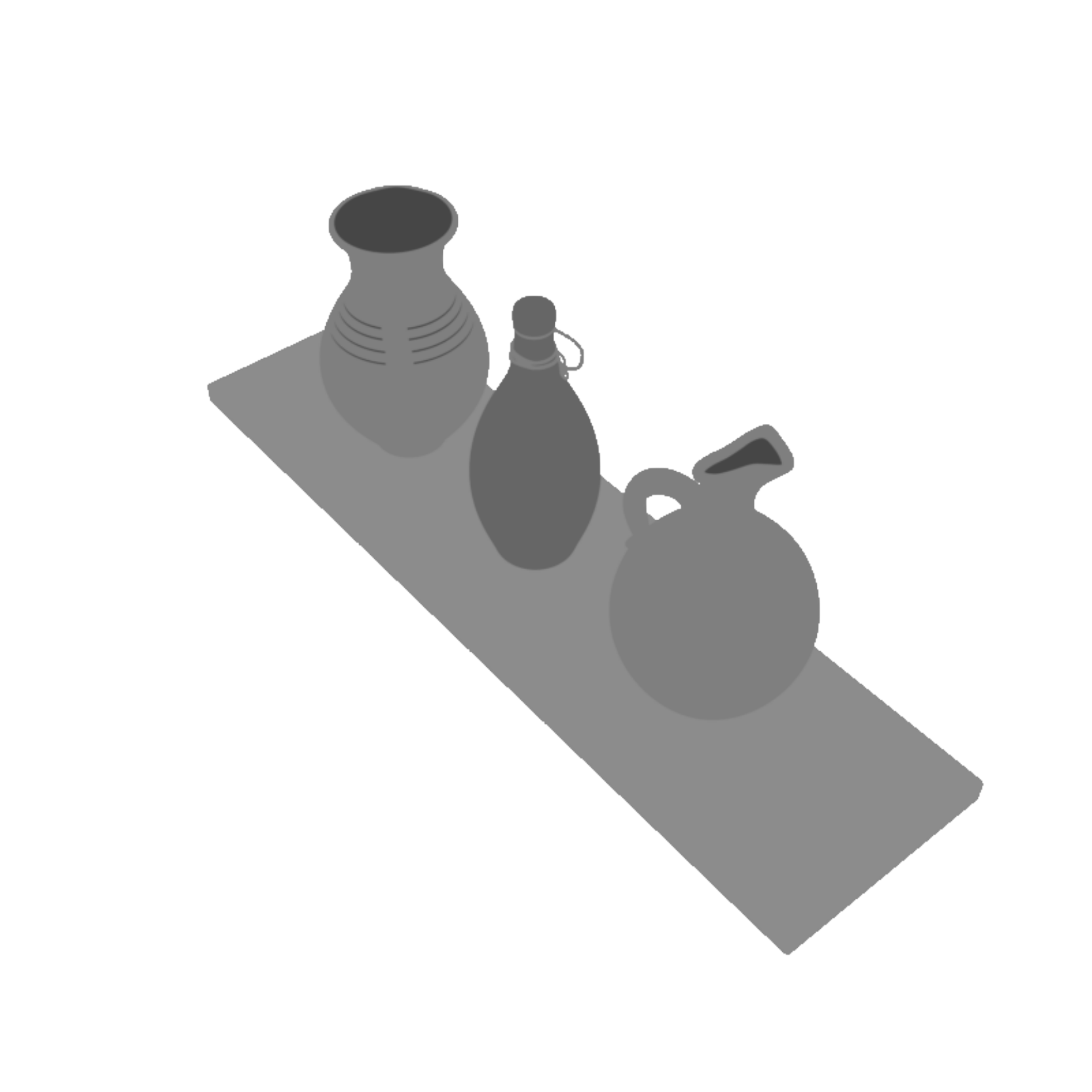} & 
\includegraphics[width=0.16\textwidth]{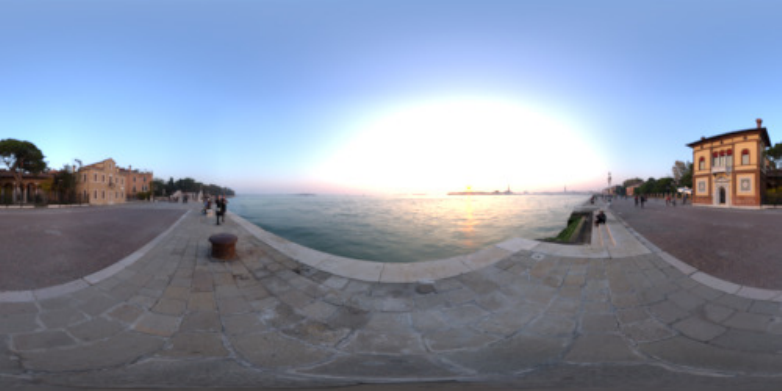} &
\includegraphics[width=0.16\textwidth]{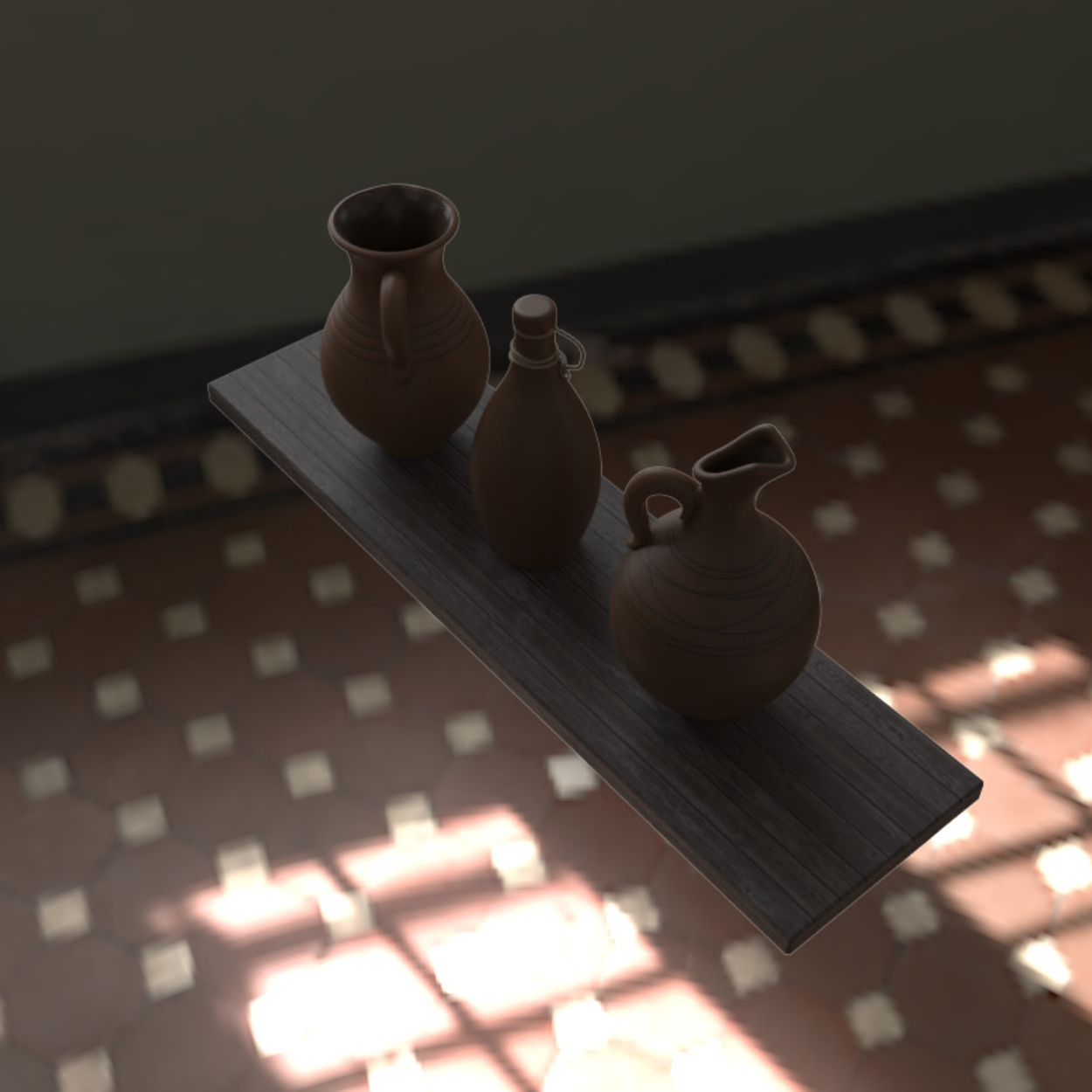} & 
\includegraphics[width=0.16\textwidth]{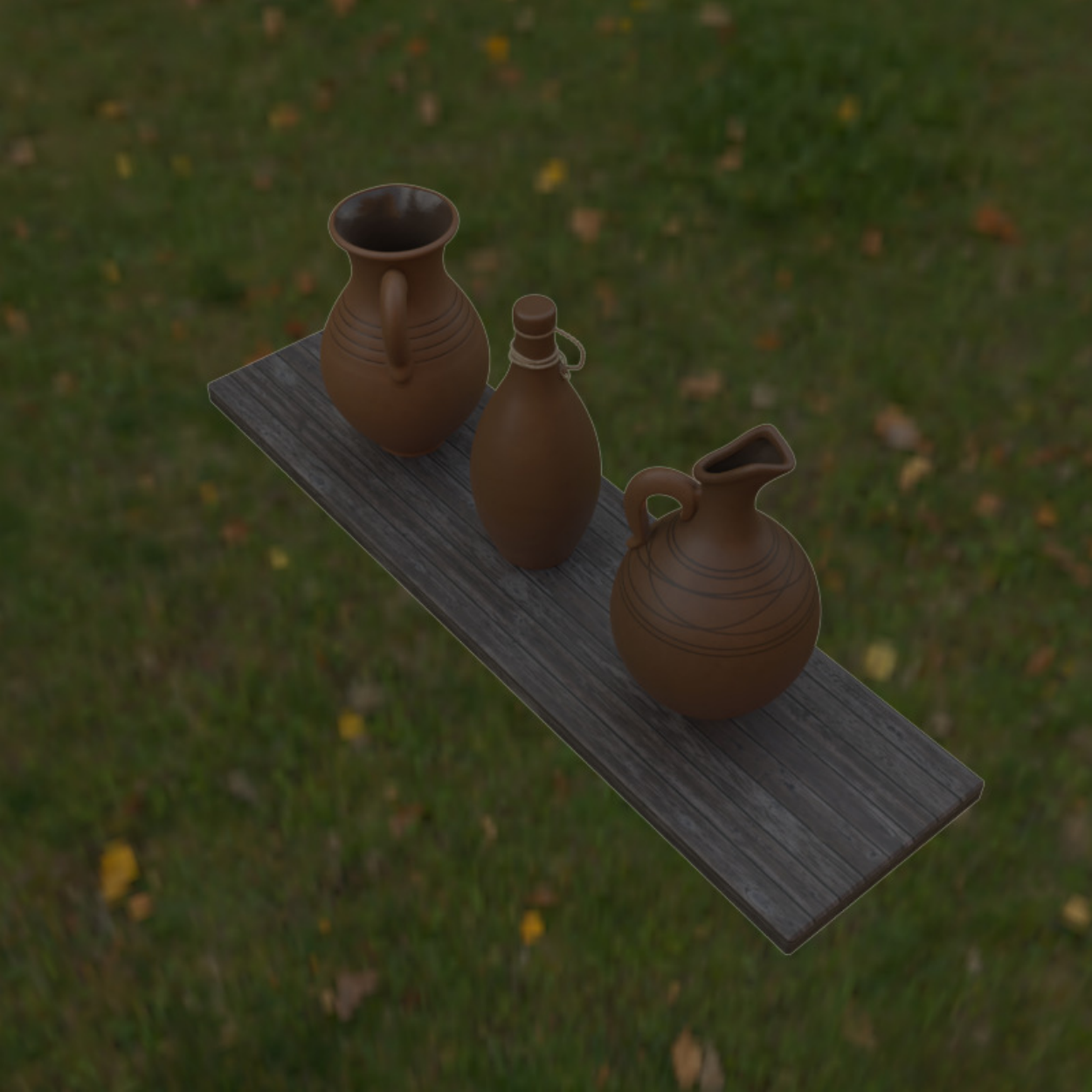} 

\end{tabular}
}
\caption{Qualitative results on Synthetic4Relight dataset~\cite{zhang2022modeling}.}
\label{fig:exp_syn4}
\end{figure}

\subsection{Performance on Relighting}
To comprehensively assess the relighting capabilities of our pipeline, we further undertook experiments using the Synthetic4Relight dataset~\cite{zhang2022modeling}. Recognizing the inherent scale ambiguity between the estimated albedo and lighting, we standardized the scale of the estimated albedo against the ground truth to relighting, consistent with previous studies~\cite{zhang2021physg,zhang2022modeling}.

Our evaluation encompasses the analysis of decomposed materials, the synthesis of novel views, and the relighting outcomes. The quantitative analysis is presented in Tab.~\ref{tab:exp_syn4}. Our method outperforms existing approaches in terms of Novel View Synthesis (NVS) and relighting precision. Regarding material estimation, our method exhibits superior albedo accuracy in SSIM and LPIPS. Qualitatively, our method achieves visually pleasing material decomposition, facilitating a realistic relighting effect, which is shown in~\ref{fig:exp_syn4}.

\begin{figure}[t]
\setlength\tabcolsep{1pt}
\centering
\resizebox{\linewidth}{!}{
\begin{tabular}{cccc}

& \raisebox{0.04\textwidth}[0pt][0pt]{Rendering} & \raisebox{0.04\textwidth}[0pt][0pt]{Relighting1} \includegraphics[height=0.08\textwidth]{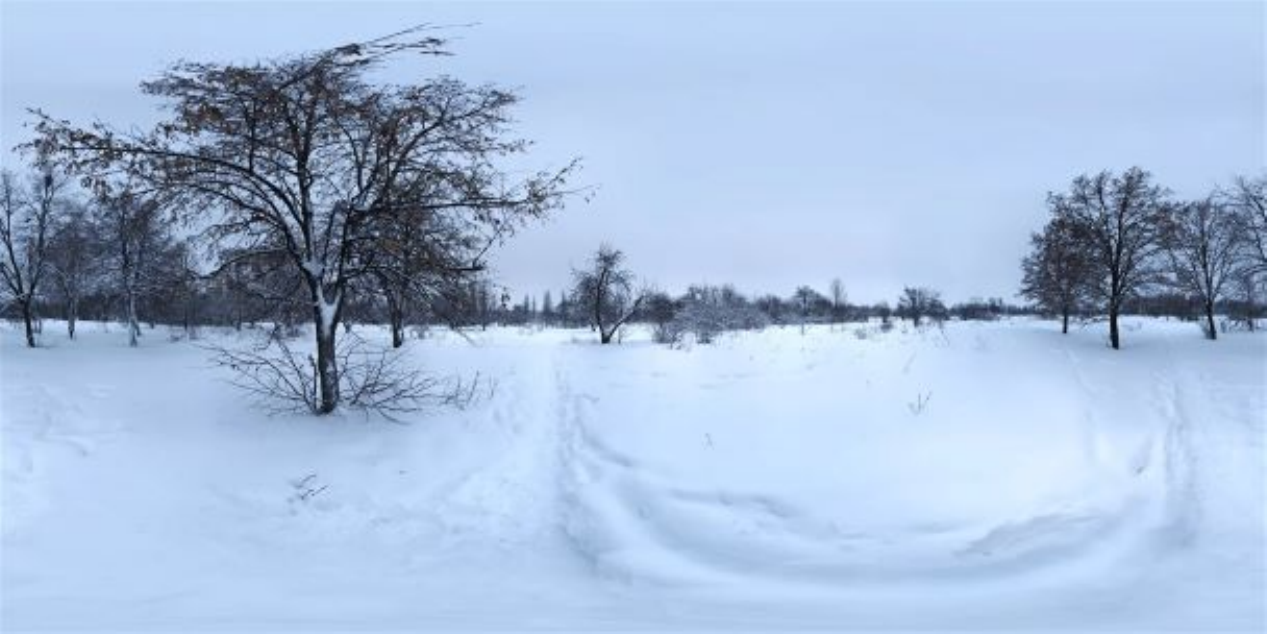} & \raisebox{0.04\textwidth}[0pt][0pt]{Relighting2} \includegraphics[height=0.08\textwidth]{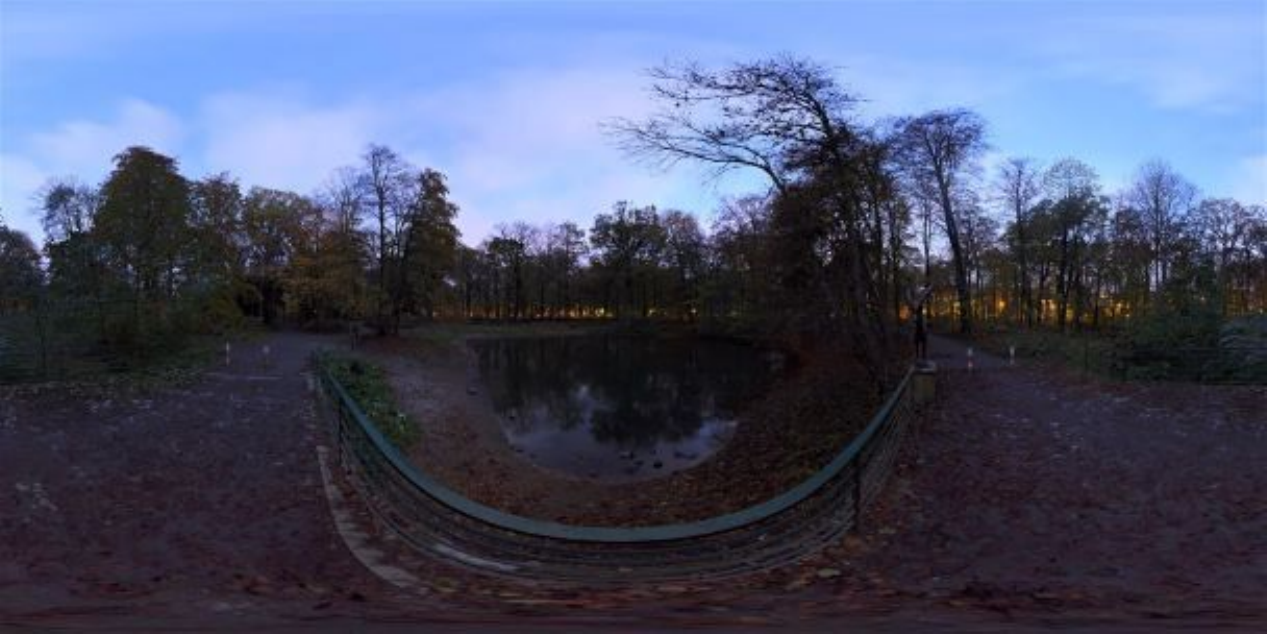}  \\
  
\raisebox{0.09\textwidth}[0pt][0pt]{\rotatebox[origin=c]{90}{Garden}} & \includegraphics[width=0.3\textwidth]{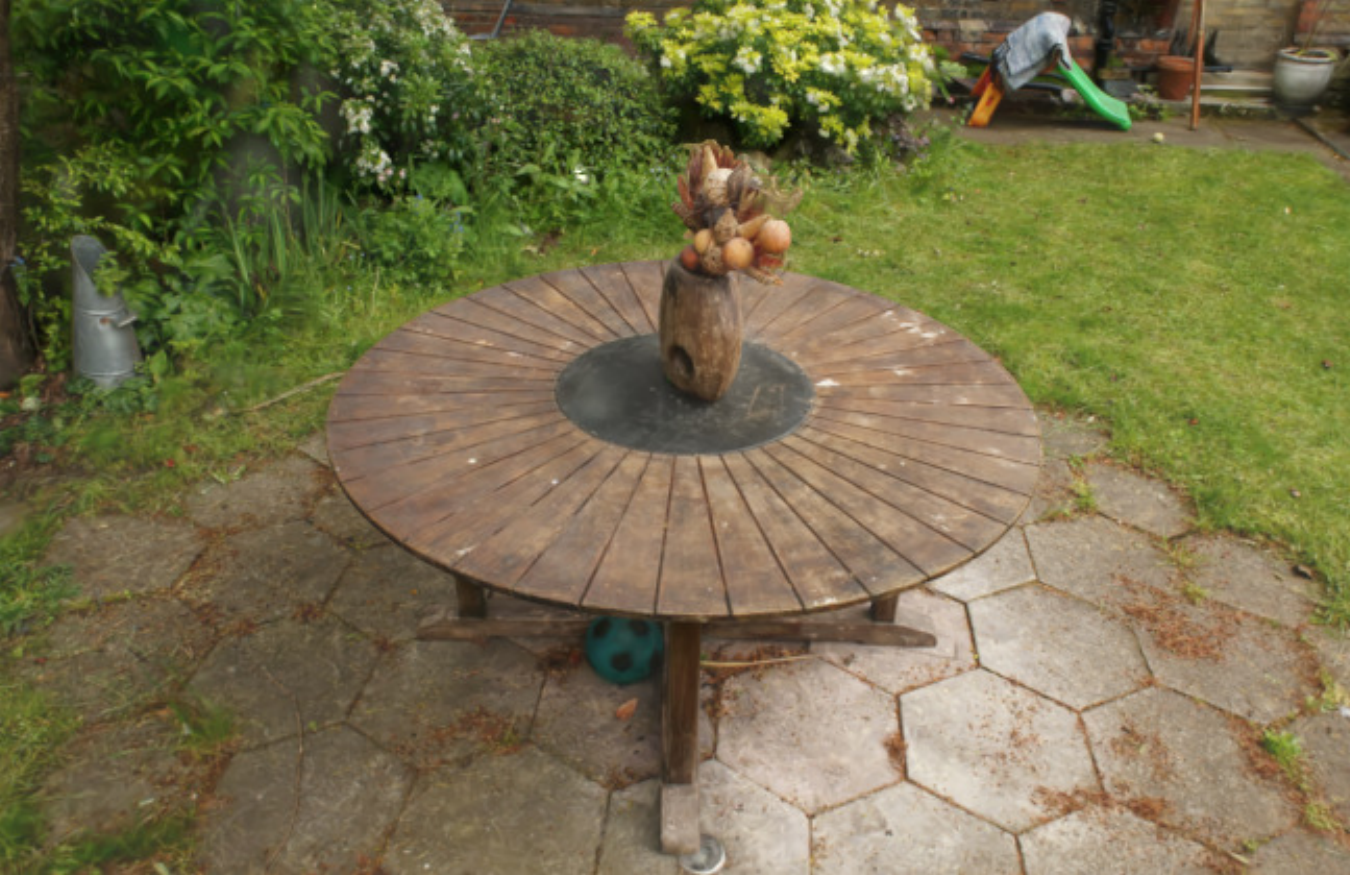} &
\includegraphics[width=0.3\textwidth]{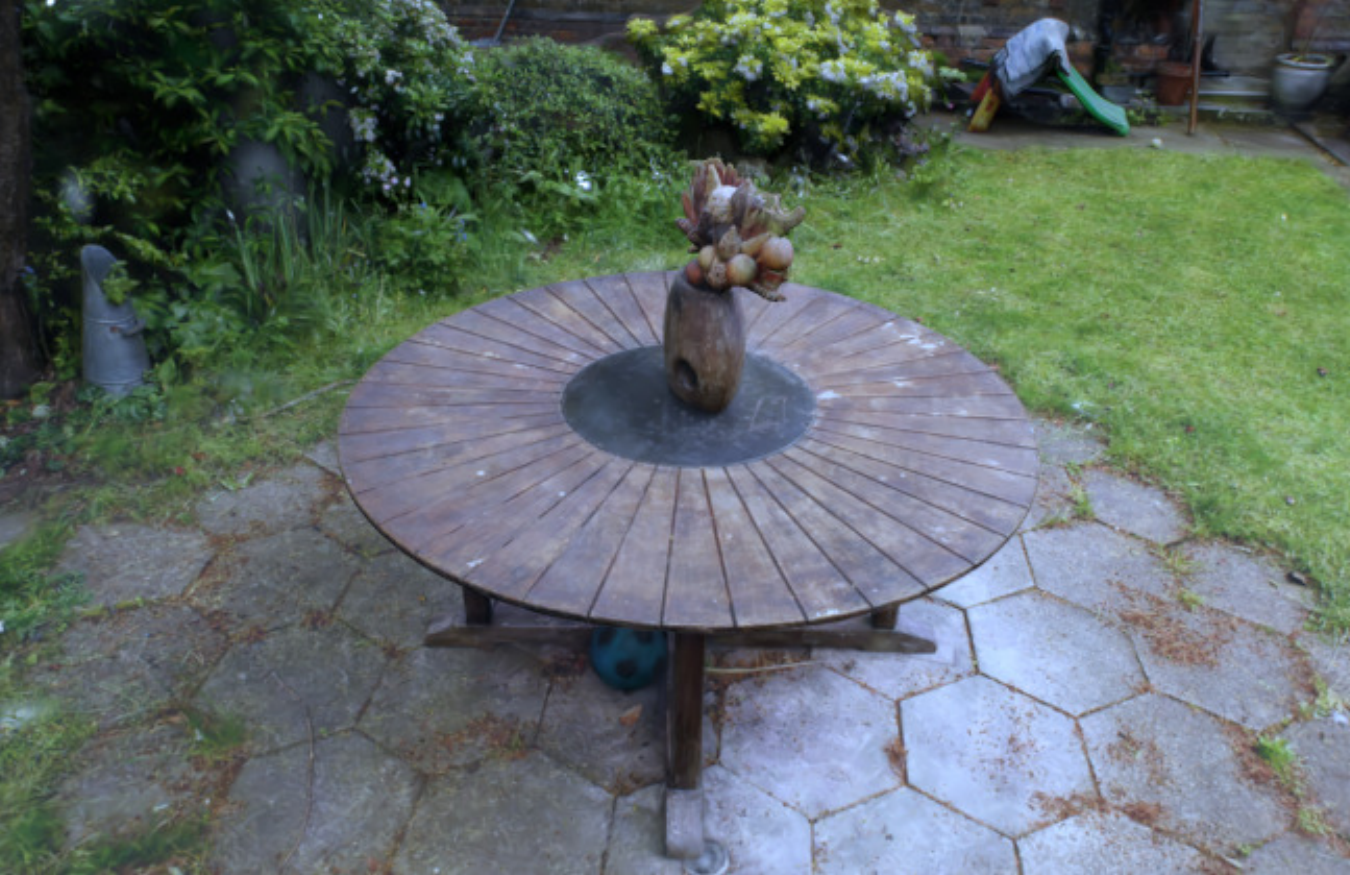} &
\includegraphics[width=0.3\textwidth]{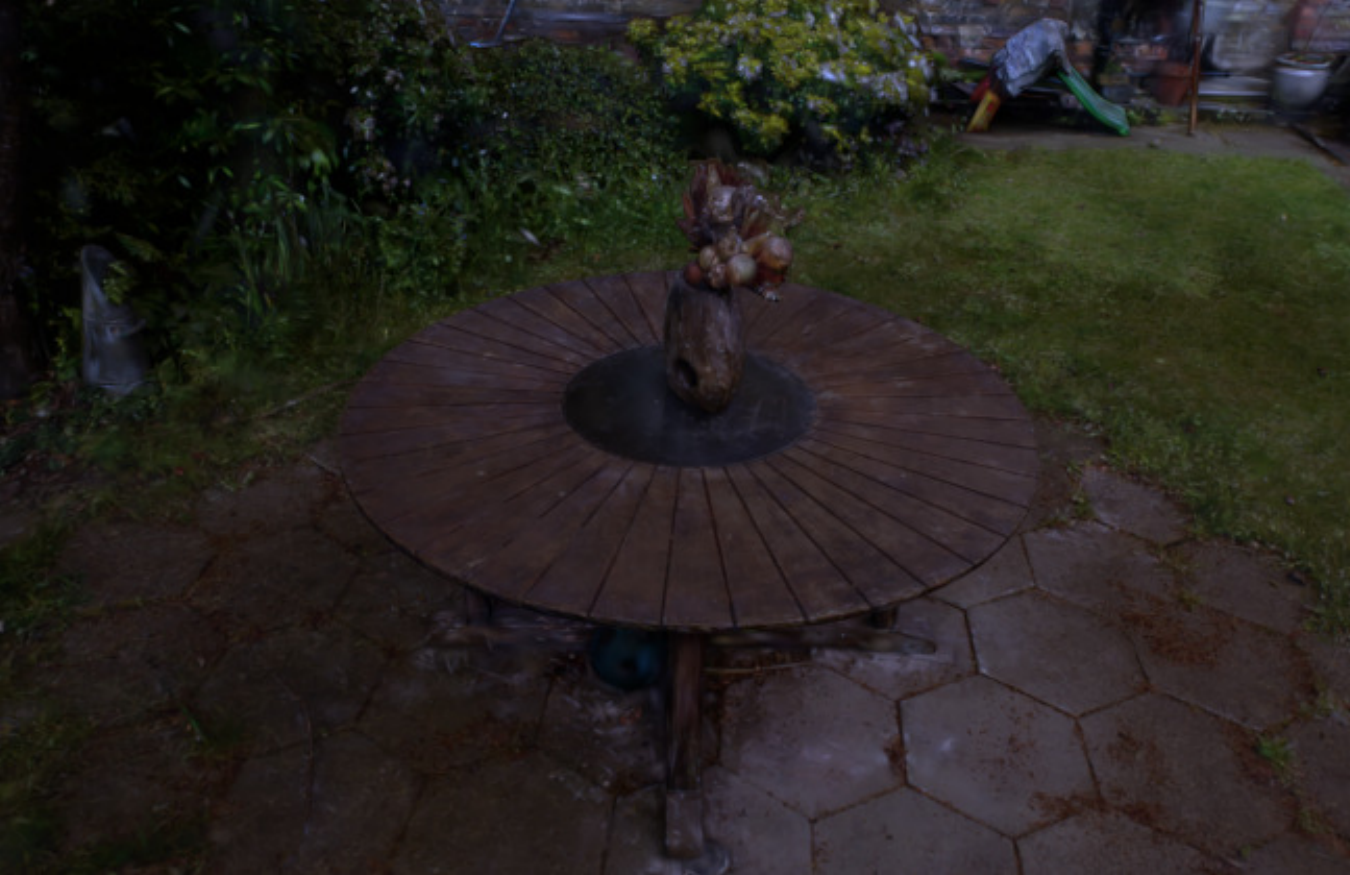} \\

\raisebox{0.09\textwidth}[0pt][0pt]{\rotatebox[origin=c]{90}{Kitchen}} & \includegraphics[width=0.3\textwidth]{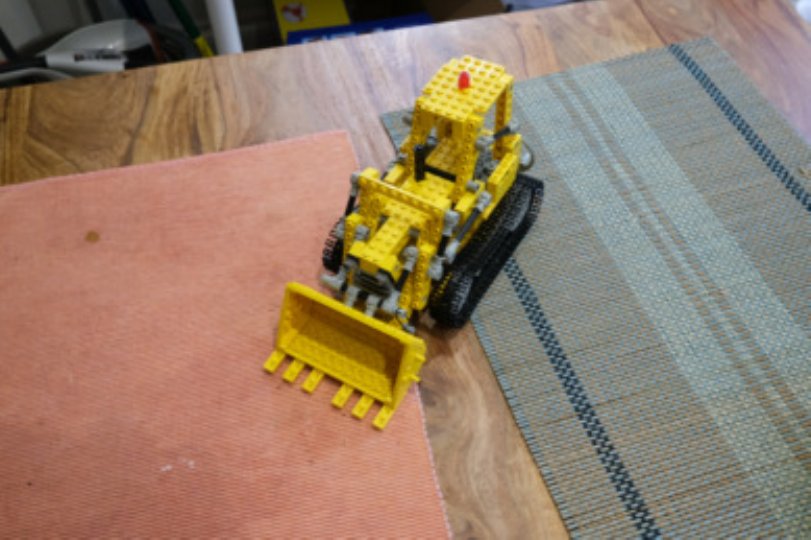} &
\includegraphics[width=0.3\textwidth]{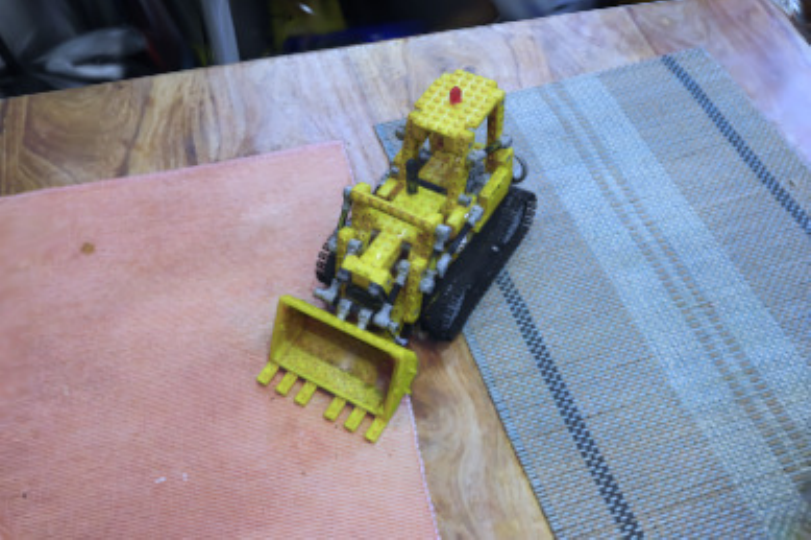} &
\includegraphics[width=0.3\textwidth]{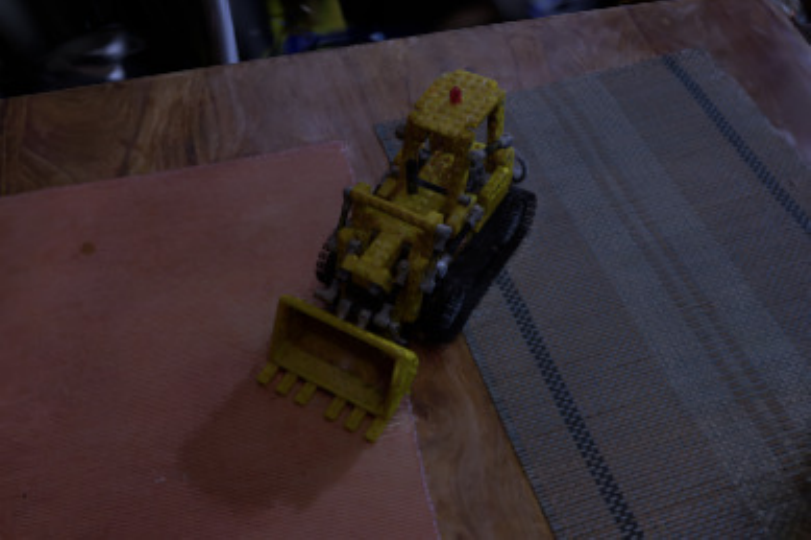} \\

\end{tabular}
}
\caption{Qualitative results of relighting on real-world scenes.}
\label{fig:scene_relighting}
\end{figure}

Additionally, we conduct relighting on real-world scenes in the Mip-NeRF 360 dataset~\cite{barron2022mip}, as shown in Fig.~\ref{fig:scene_relighting}. It is worth noting that our relighting in the \textit{kitchen} with the second environment map, exhibits pleasing shadow effects.

\subsection{Ablation Study}
\label{sec:exp_ablation}

\begin{figure}[tb]
    \begin{subfigure}[b]{0.19\linewidth}
        \includegraphics[width=\linewidth]{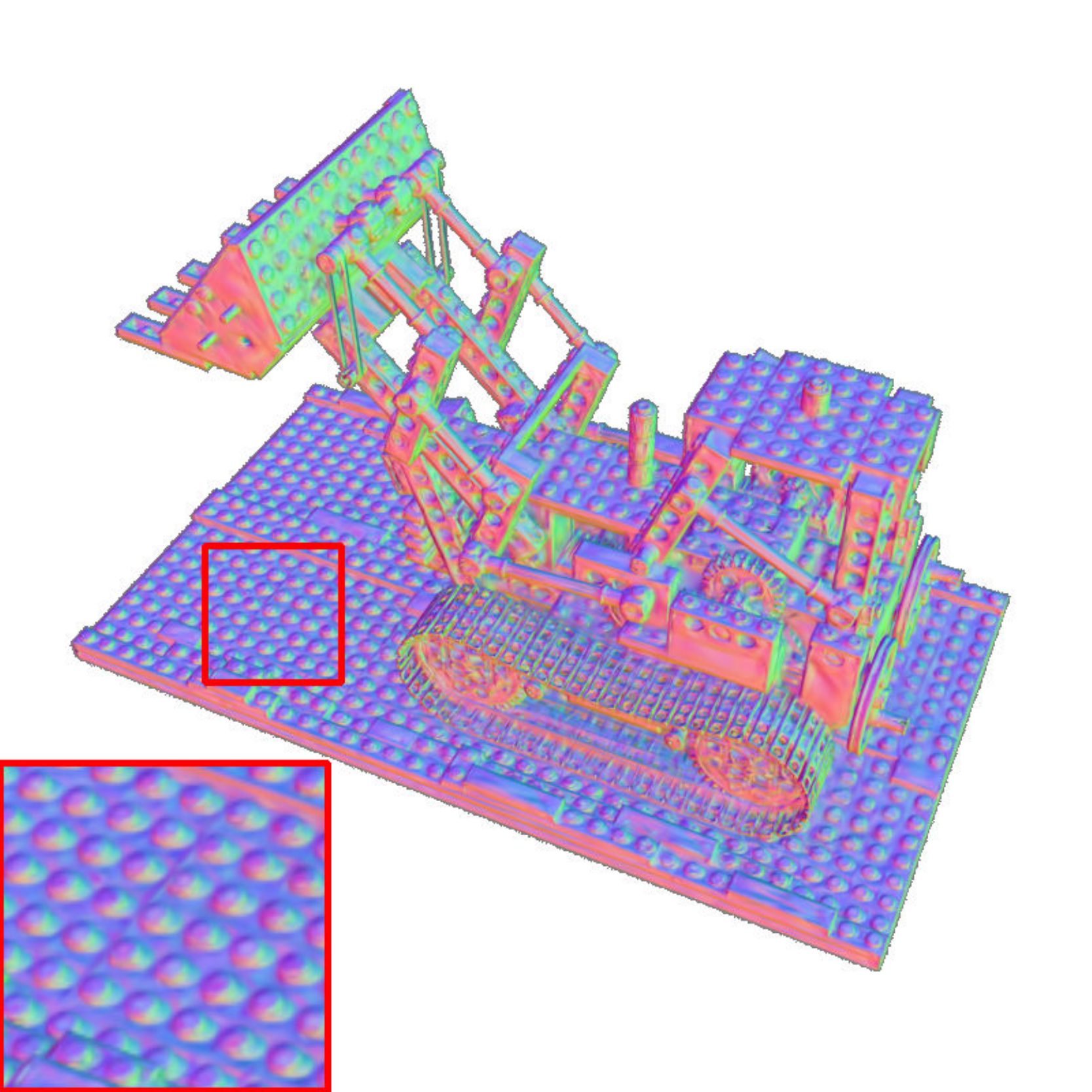}
        \caption{w NGD.}
        \label{fig:ablation_ngd1}
    \end{subfigure}
    \begin{subfigure}[b]{0.19\linewidth}
        \includegraphics[width=\linewidth]{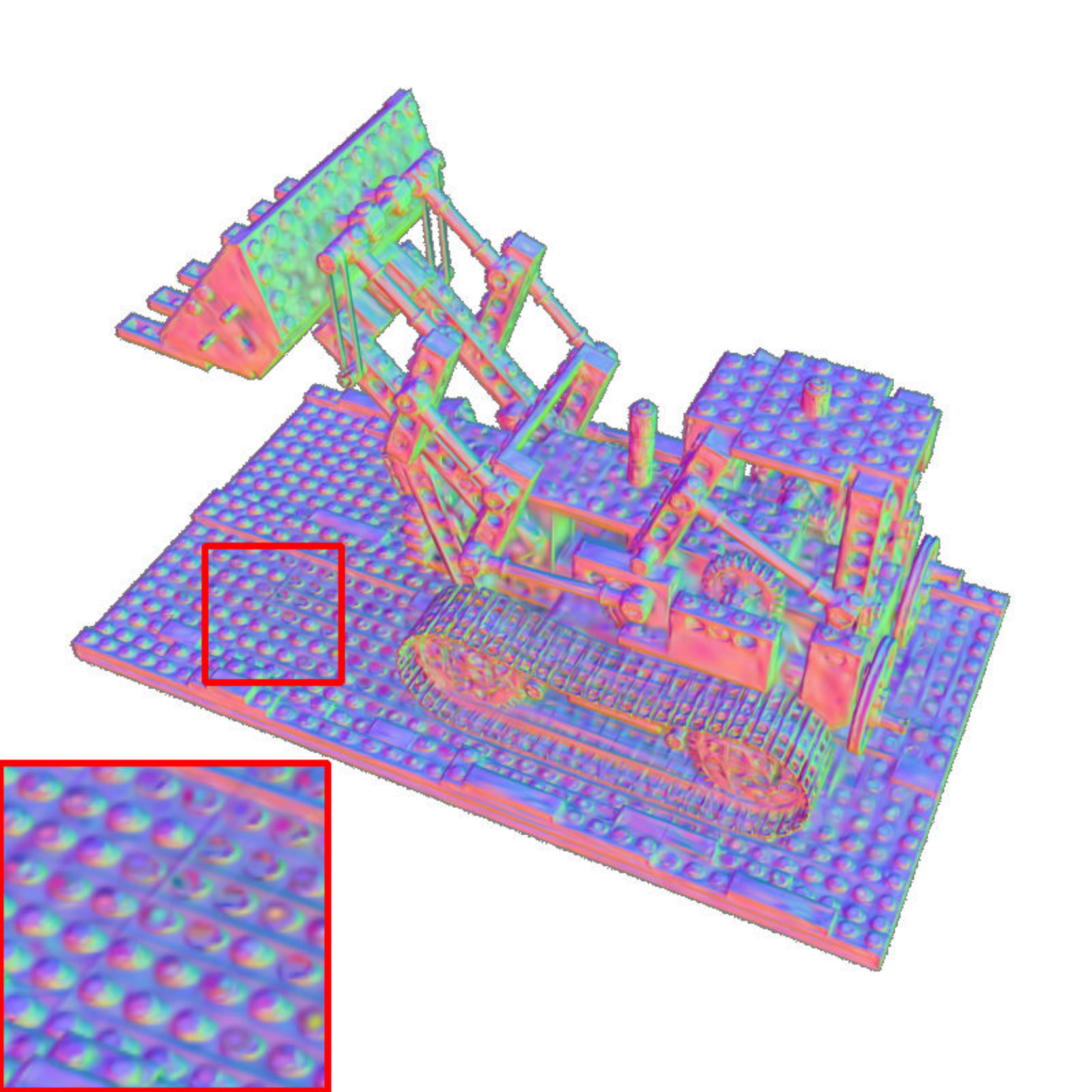}
        \caption{w/o NGD.}
        \label{fig:ablation_ngd2}
    \end{subfigure}
    \begin{subfigure}[b]{0.19\linewidth}
        \includegraphics[width=\linewidth]{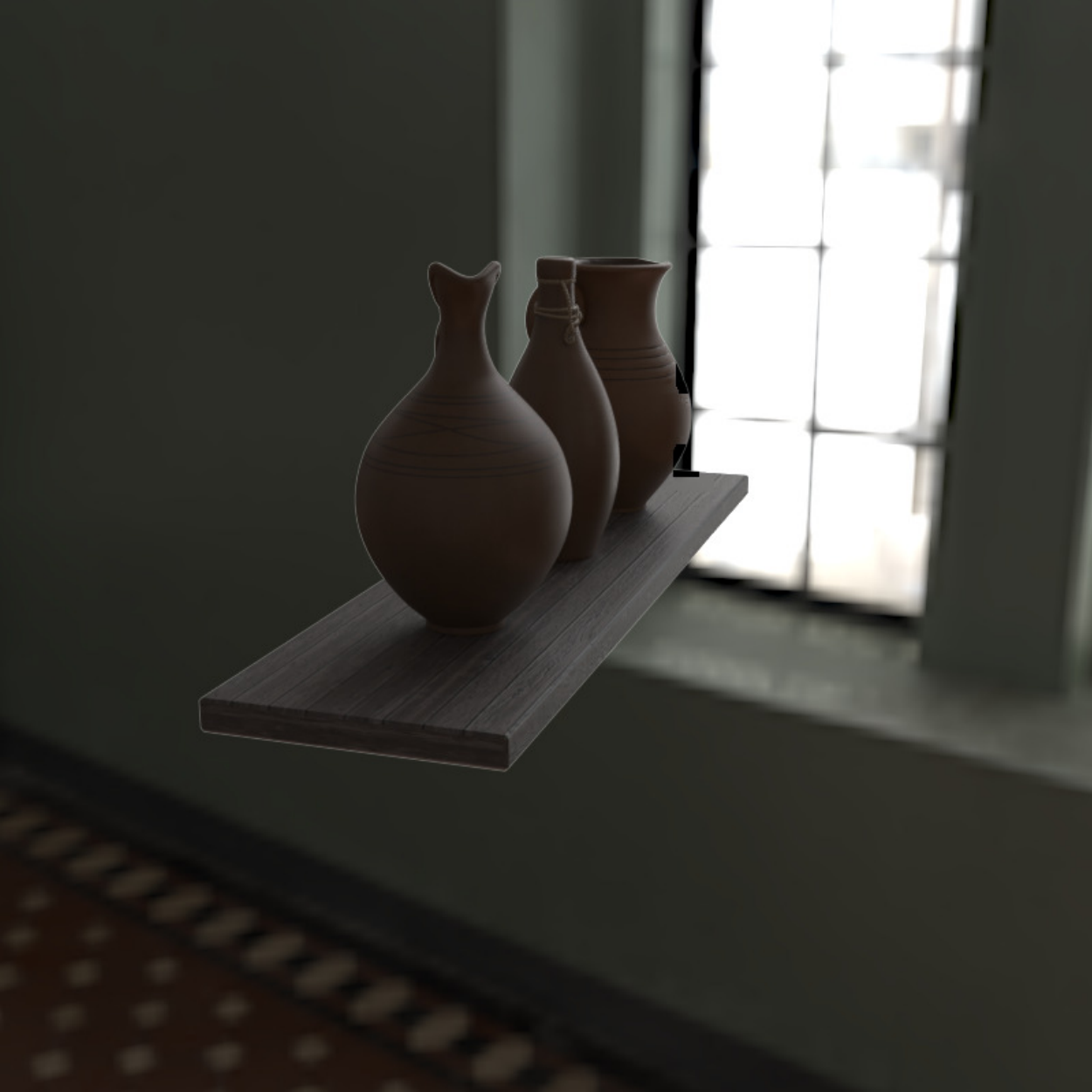}
        \caption{Relit. GT. }
        \label{fig:ablation_relit1}
    \end{subfigure}
    \begin{subfigure}[b]{0.19\linewidth}
        \includegraphics[width=\linewidth]{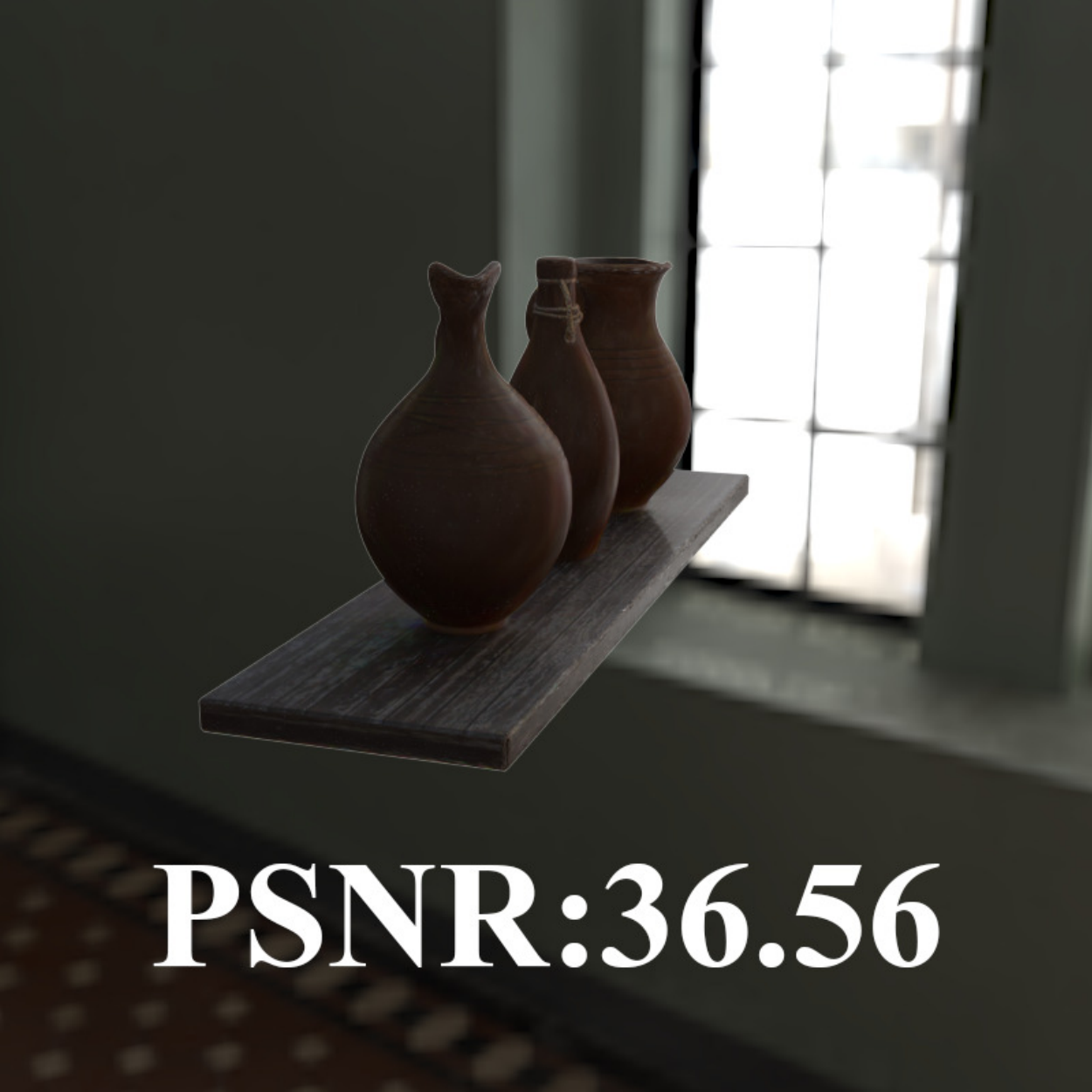}
        \caption{Full light. }
        \label{fig:ablation_relit2}
    \end{subfigure}
    \begin{subfigure}[b]{0.19\linewidth}
        \includegraphics[width=\linewidth]{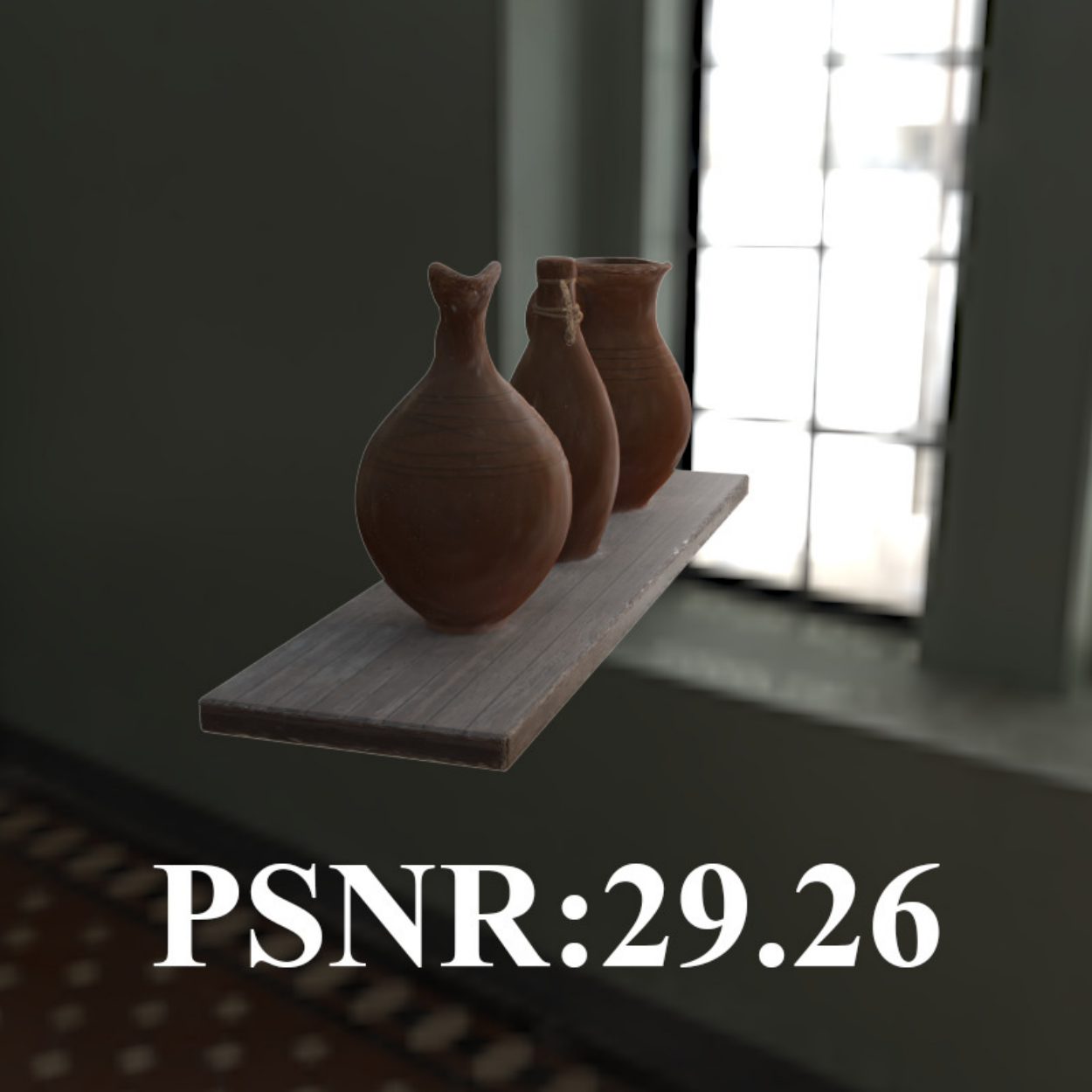}
        \caption{Simplified. }
        \label{fig:ablation_relit3}
    \end{subfigure}
    
    \begin{subfigure}[b]{0.24\linewidth}
        \includegraphics[width=\linewidth]{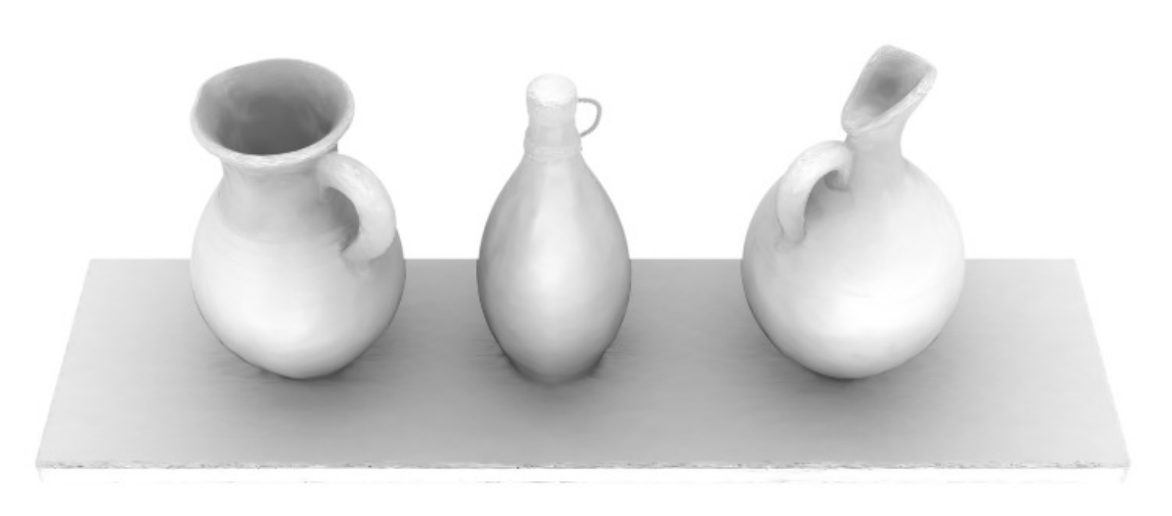}
        \caption{AO w $\mathcal{L}_{u}$. }
        \label{fig:ablation_var1}
    \end{subfigure}
    \begin{subfigure}[b]{0.24\linewidth}
        \includegraphics[width=\linewidth]{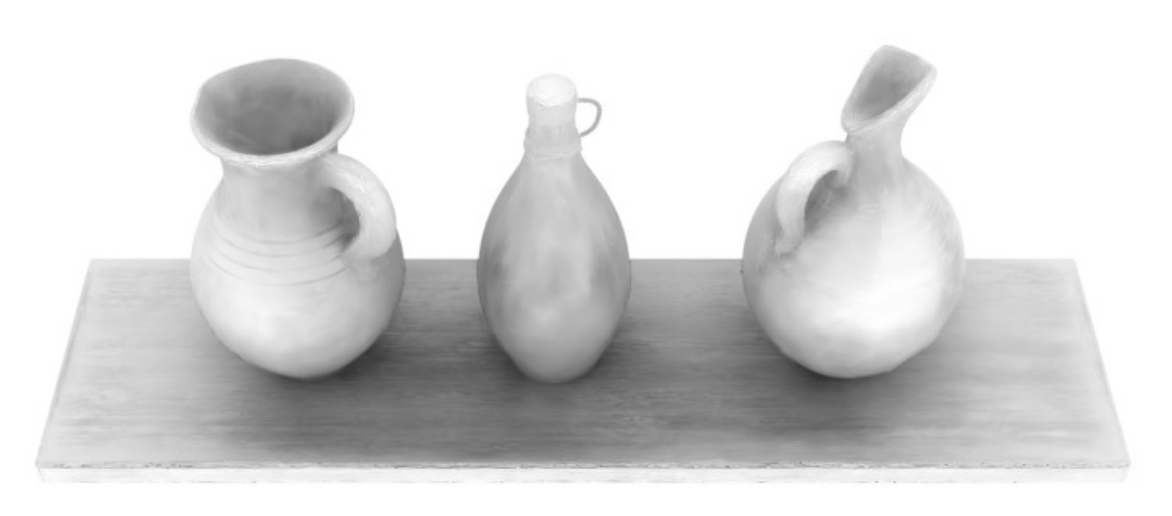}
        \caption{AO w/o $\mathcal{L}_{u}$. }
        \label{fig:ablation_var2}
    \end{subfigure}
    \begin{subfigure}[b]{0.24\linewidth}
        \includegraphics[width=\linewidth]{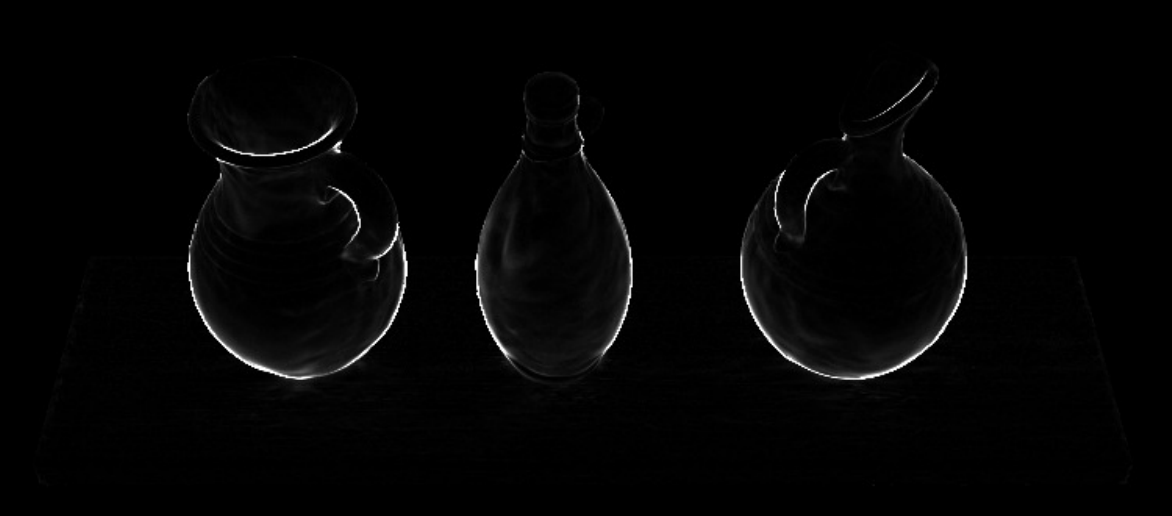}
        \caption{Uncertainty w $\mathcal{L}_{u}$. }
        \label{fig:ablation_var3}
    \end{subfigure}
    \begin{subfigure}[b]{0.24\linewidth}
        \includegraphics[width=\linewidth]{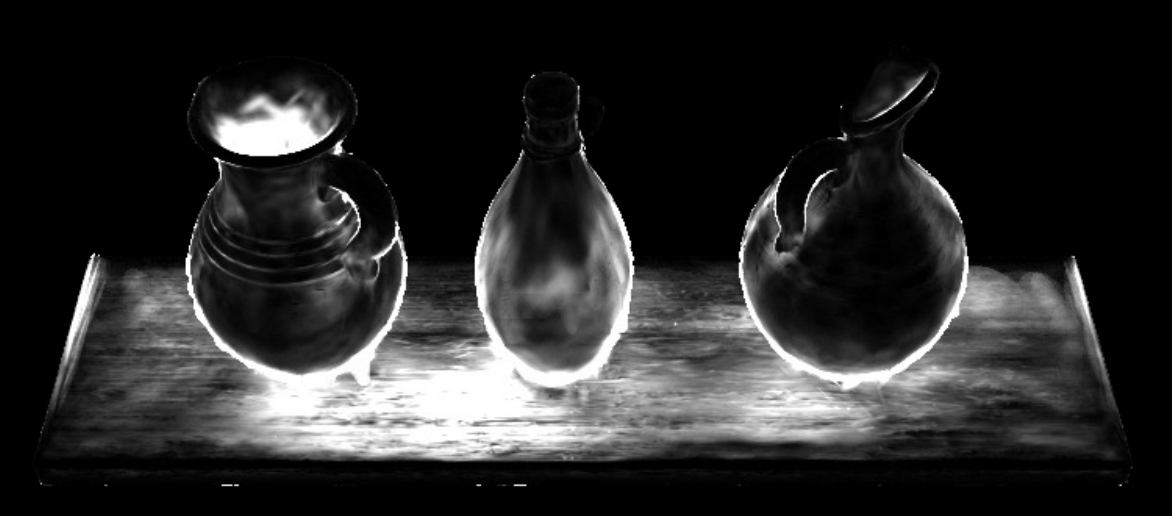}
        \caption{Uncertainty w/o $\mathcal{L}_{u}$. }
        \label{fig:ablation_var4}
    \end{subfigure}
    \caption{\textbf{Ablation studies on main components of our method.} (a) and (b) show normal maps with and without the proposed normal gradient densification (NGD). (d) and (e) display re-lighting results using the proposed full lighting model and only a simplified light modeling, i.e. a global environment map. (f) and (g) visualize ambient occlusion (AO) maps with and without the proposed constraint on depth distribution $\mathcal{L}_{u}$. (h) and (i) illustrate the depth uncertainty with and without $\mathcal{L}_{u}$.}
    \label{fig:ablation}
\end{figure}

We conduct ablation studies on three principal components of our method that significantly influence the quality of inverse rendering. The results are illustrated in Fig.~\ref{fig:ablation}. Initially, we investigate the impact of normal gradient densification (NGD). Fig.~\ref{fig:ablation} (a, b) depict the rendering normal maps with and without the densification. It shows that the proposed densification markedly enhances the details of the normal map for \textit{Lego}. In Fig.~\ref{fig:ablation} (c, d, e), we examine the effects of our full light modeling. We simplify the full model to a global environment map in the optimization. The result demonstrates that the simplified model fails to generate shadow effects when relighting. Furthermore, we evaluate the significance of the proposed constraint on depth distribution, denoted as $\mathcal{L}_{u}$. The depth uncertainty is visualized in Fig.~\ref{fig:ablation} (h, i), which illustrates the distribution constraint contribute significantly to reducing the depth uncertainty. Additionally, Fig.~\ref{fig:ablation} (f, g) reveals that incorporating $\mathcal{L}_{u}$ leads to more plausible ambient occlusion (averaged visibility) which plays a crucial role in stage 2.

\section{Conclusion}
\label{sec:conclusion}
In this paper, we have introduced a differentiable point-based rendering pipeline enabling effort less editing, accuracy ray tracing, and photo-realistic relighting. We present the scene as \textit{Relightable 3D Gaussians}, an extension of traditional 3D Gaussians enriched with supplementary normals, BRDFs and indirect lighting. To reconstruct a relightable scene from multi-view images, we build a novel differentiable rendering pipeline based on the inverse rendering techniques and 3D Gaussian Splatting. Additionally, we introduce an innovative ray tracing scheme specifically designed for scenes represented by discrete points, providing precise visibility for realistic relighting. Quantitative and qualitative experiments confirm that our pipeline successfully reconstructs reasonable normals, materials for discrete point clouds, and achieves commendable accuracy in both novel view synthesis and scene relighting tasks.

\noindent{\bf Limitations and Future work.} Our present pipeline targets the reconstruction of static objects. Certain design choices pose challenges in maintaining its performance with large scale scenes. The high density of points in large scene can slow down optimization as we sample rays and perform PBR at each point. This can be mitigated through deferred rendering technique. Regarding geometry, integrating MVS shows promise for achieving more accurate representations.

\section*{Acknowledgements} This work was supported by National Key R\&D Program of China (2023YFB3209702), and NSFC (62441204).

%
%
\bibliographystyle{splncs04}
\bibliography{main}
\end{document}